\documentclass[10pt,twocolumn,letterpaper]{article}

\usepackage{cvpr}              
\usepackage[accsupp]{axessibility}  
\usepackage{graphicx}
\usepackage{amsmath}
\usepackage{amssymb}
\usepackage{booktabs}
\usepackage{multirow}
\usepackage{soul}
\usepackage[table,dvipsnames]{xcolor}

\usepackage[pagebackref,breaklinks,colorlinks]{hyperref}

\usepackage[capitalize]{cleveref}
\crefname{section}{Sec.}{Secs.}
\Crefname{section}{Section}{Sections}
\Crefname{table}{Table}{Tables}
\crefname{table}{Tab.}{Tabs.}

\def\cvprPaperID{3553} 
\def\confName{CVPR}
\def\confYear{2023}

\definecolor{yellow}{rgb}{1,0.97, 0.65}
\definecolor{lightyellow}{rgb}{1,1, 0.8}
\definecolor{orange}{rgb}{1, 0.85, 0.7}
\definecolor{tablered}{rgb}{1, 0.7, 0.7}

\newcommand{\methodname}{TensoIR}

\newcommand{\boldstart}[1]{\noindent\textbf{#1}}
\newcommand{\boldstartspace}[1]{\vspace{0.1in}\noindent\textbf{#1}}

\newcommand\blfootnote[1]{%
  \begingroup
  \renewcommand\thefootnote{}\footnote{#1}%
  \addtocounter{footnote}{-1}%
  \endgroup
}

\begin{document}

\title{\methodname: Tensorial Inverse Rendering}
\vspace{-3mm}
\author{
Haian Jin$^{*1}$ \hspace{0.5cm} 
Isabella Liu$^{*2}$ \hspace{0.5cm} 
Peijia Xu$^{3}$ \hspace{0.5cm} 
Xiaoshuai Zhang$^2$ \hspace{0.5cm} 
Songfang Han$^2$
\\
Sai Bi$^4$ \hspace{0.5cm} 
Xiaowei Zhou$^1$ \hspace{0.5cm} 
Zexiang Xu$^{\textsuperscript{\textdagger} 4}$ \hspace{0.5cm}
Hao Su$^{\textsuperscript{\textdagger} 2}$
\\
\\
$^1$ Zhejiang University \hspace{0.35cm} 
$^2$ UC San Diego \hspace{0.35cm}
$^3$ Kingstar Technology Inc. \hspace{0.35cm} 
$^4$ Adobe Research
}
\vspace{-3mm}
\maketitle

\blfootnote{Project page: \url{https://haian-jin.github.io/TensoIR}}
\blfootnote{* Equal contribution. \textsuperscript{\textdagger} Equal advisory.}

\begin{abstract}
    We propose \methodname, a novel inverse rendering approach based on tensor factorization and neural fields. 
    Unlike previous works that use purely MLP-based neural fields, thus suffering from low capacity and high computation costs, we extend TensoRF, a state-of-the-art approach for radiance field modeling, to estimate scene geometry, surface reflectance, and environment illumination from multi-view images captured under unknown lighting conditions. 
    Our approach jointly achieves radiance field reconstruction and physically-based model estimation, leading to photo-realistic novel view synthesis and relighting results. 
    Benefiting from the efficiency and extensibility of the TensoRF-based representation, our method can accurately model secondary shading effects (like shadows and indirect lighting) and generally support input images captured under single or multiple unknown lighting conditions. The low-rank tensor representation allows us to not only achieve fast and compact reconstruction but also better exploit shared information under an arbitrary number of capturing lighting conditions.
    We demonstrate the superiority of our method to baseline methods qualitatively and quantitatively on various challenging synthetic and real-world scenes.
\end{abstract}
\section{Introduction}
\label{sec:intro}

Inverse rendering is a long-standing problem in computer vision and graphics,
aiming to reconstruct physical attributes (like shape and materials) of a 3D
scene from captured images and thereby supporting many downstream applications such as novel view synthesis, relighting
and material editing.
This problem is inherently challenging and ill-posed, especially when the input
images are captured in the wild under unknown illumination.
Recent works~\cite{boss2021nerd,zhang2021nerfactor, NeuralPIL, srinivasan2021nerv} address this problem by learning
neural field representations in the form of multi-layer perceptrons (MLP) similar to NeRF~\cite{mildenhall2021nerf}.
However, pure MLP-based methods usually suffer from low capacity and high computational costs, greatly limiting the accuracy and efficiency of inverse rendering.

\begin{figure}[t]
\includegraphics[width=1\linewidth]{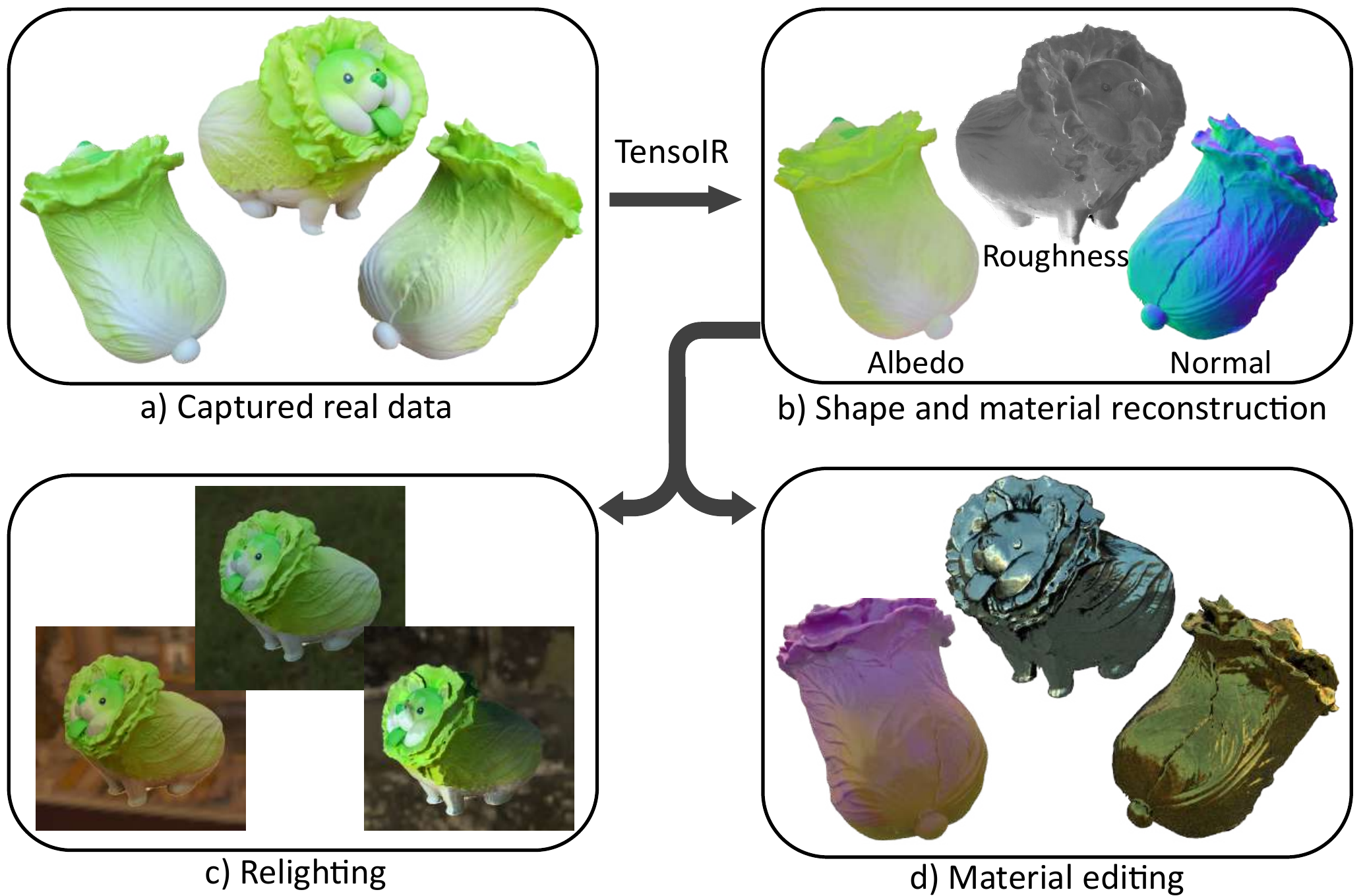}
    \caption{Given multi-view captured images of a real scene (a), our approach -- TensoIR -- is able to achieve high-quality shape and material reconstruction with high-frequency details (b). This allows us to render the scene under novel lighting and viewpoints (c), and also change its material properties (d).}
    \label{fig:teaser}
    \vspace{-1mm}
\end{figure}

In this work, we propose a novel inverse rendering framework that is efficient and accurate. 
Instead of purely using MLPs, we build upon the recent 
TensoRF~\cite{chen2022tensorf} scene representation, which achieves fast, compact, and state-of-the-art quality on radiance field reconstruction for novel
view synthesis. Our tensor factorization-based inverse rendering framework can 
simultaneously estimate scene geometry, materials, and illumination from
multi-view images captured under unknown lighting conditions.
Benefiting from the efficiency and extensibility of the TensoRF representation, our method 
can accurately model secondary shading effects (like shadows and indirect lighting) and generally support input images captured under a single or multiple unknown lighting conditions.

Similar to TensoRF, our approach models a scene as a neural voxel feature grid, factorized as multiple low-rank tensor
components. 
We apply multiple small MLPs on the same feature grid 
and regress volume density, view-dependent color, normal, and material properties, to model the scene geometry and appearance.
This allows us to simultaneously achieve both radiance field rendering -- using density and view-dependent color, as done in NeRF \cite{mildenhall2021nerf} -- and physically-based rendering -- using density, normal and material properties, as done in inverse rendering methods \cite{li2018learning,bi2020neural}.
We supervise both renderings with the captured images 
to jointly reconstruct all scene components.
In essence, we reconstruct a scene using both a radiance field and a physically-based model to reproduce the scene's appearance.
While inverse rendering is our focus and primarily enabled by the physically-based model, modeling the radiance field is crucial for the success of the reconstruction (see Fig.~\ref{fig:rfcompare}), in significantly facilitating the volume density reconstruction and effectively regularizing the same tensor features shared by the physically-based model.
Despite that previous works~\cite{zhang2021nerfactor} similarly reconstruct NeRFs in inverse rendering, their radiance field is pre-computed and fixed in the subsequent inverse rendering stage; in contrast, our radiance field is jointly reconstructed and also benefits the physically-based rendering model estimation during optimization, leading to much higher quality.
Besides, our radiance field rendering can also be directly used to provide accurate indirect illumination for the physically-based rendering, further benefiting the inverse rendering process.

Accounting for indirect illumination and shadowing is a critical challenge 
in inverse rendering.
This is especially challenging for volume rendering, since it requires sampling a lot of secondary rays and computing the 
integrals along the rays by performing ray marching.

Limited by the high-cost MLP
evaluation, previous NeRF-based methods and SDF-based methods either simply ignore secondary effects~\cite{boss2021nerd,NeuralPIL,zhang2021physg}, or avoid online computation by approximating these effects in extra distilled MLPs~\cite{zhang2021nerfactor,InvRenderer}, requiring expensive pre-computation and leading to degradation in accuracy.

In contrast, owing to our efficient tensor-factorized representation, 
we are able to explicitly compute the ray integrals online for
accurate visibility and indirect lighting with the radiance field rendering using low-cost second-bounce ray marching.
Consequently, our approach enables higher accuracy in modeling
these secondary effects, which is crucial in achieving 
high-quality scene reconstruction (see Tab.~\ref{tab:ablation_indirect_light}). 

In addition, the flexibility and efficiency of our tensor-factorized
representation allows us to perform inverse rendering from multiple   
unknown lighting conditions with limited GPU memory.  
Multi-light capture is known to be beneficial for inverse rendering tasks by providing useful photometric cues and reducing ambiguities in material estimation,
thus being commonly used~\cite{debevec2000acquiring,lensch2003ligtstage,goldman2009shape}.
However, since each lighting condition corresponds to a separate radiance field, this can lead to extremely high computational costs if reconstructing multiple purely MLP-based NeRFs like previous works~\cite{zhang2021nerfactor,InvRenderer, srinivasan2021nerv}.
Instead, we propose to reconstruct radiance fields under multi-light in a 
joint manner as a factorized tensor. 
Extending from the original TensoRF representation that is a 4D tensor, we add an additional dimension corresponding to different lighting conditions, yielding a 5D tensor. 
Specifically, we add an additional vector factor (whose length equals the number
of lights) per tensor component to explain the appearance variations under different lighting conditions, and we store this 5D tensor by a small number of bases whose outer-product reconstructs the 5D tensor.
When multi-light capture is available, our framework can effectively
utilize the additional photometric cues in the data, leading to better
reconstruction quality than a single-light setting (see Tab.~\ref{tab:main_res}).

As shown in Fig.~\ref{fig:teaser}, our approach can reconstruct high-fidelity geometry and reflectance of a complex real scene captured under unknown natural illumination, enabling photo-realistic rendering under novel lighting conditions and additional applications like material editing. 
We evaluate our framework extensively on both synthetic and real data.
Our approach outperforms previous inverse rendering methods~\cite{zhang2021nerfactor,InvRenderer} by a large margin qualitatively and quantitatively on challenging synthetic scenes, achieving state-of-the-art quality in scene reconstruction -- for both geometry and material properties --  and rendering -- for both novel view synthesis and relighting.
Owing to our efficient tensorial representation and joint reconstruction scheme, our approach also leads to a much faster reconstruction speed than previous neural field-based reconstruction methods while achieving superior quality.
In summary,
\begin{itemize}
\vspace{-1mm}
\item We propose a novel tensor factorization-based inverse rendering approach that jointly achieves physically-based rendering model estimation and radiance field reconstruction, leading to state-of-the-art scene reconstruction results;
\vspace{-2mm}
\item Our framework includes an efficient online visibility and indirect lighting computation technique, providing accurate second-bounce shading effects;
\vspace{-2mm}
\item We enable efficient multi-light reconstruction by modeling an additional lighting dimension in the factorized tensorial representation. 

\end{itemize}

\newcommand{\Camorg}{\mathbf{o}}
\newcommand{\Raydir}{\mathbf{d}}
\newcommand{\Ray}{\mathbf{r}}

\newcommand{\Dens}{\sigma}
\newcommand{\Rad}{c}
\newcommand{\VolW}{w}
\newcommand{\Trans}{T}
\newcommand{\iiRay}{j}
\newcommand{\iiTran}{q}
\newcommand{\iiLight}{l}
\newcommand{\iiRank}{k}

\newcommand{\Color}{C}
\newcommand{\ColorGT}{C_{\text{gt}}}
\newcommand{\ColorRF}{C_{\text{RF}}}
\newcommand{\ColorPB}{C_{\text{PB}}}

\newcommand{\Pos}{\mathbf{x}}
\newcommand{\SurfP}{\mathbf{\hat{x}}}
\newcommand{\SurfDirO}{\Raydir}
\newcommand{\SurfDirIn}{\boldsymbol{\omega}_i}

\newcommand{\Lin}{L_{\text{i}}}
\newcommand{\Ldirect}{L_{\text{d}}}
\newcommand{\Lindi}{L_{\text{ind}}}
\newcommand{\Vis}{V}

\newcommand{\Normal}{\mathbf{n}}
\newcommand{\Albedo}{a}
\newcommand{\Rough}{\rho}
\newcommand{\BRDFunc}{f_r}
\newcommand{\BRDFParam}{\boldsymbol{\beta}}

\newcommand{\MLP}{\mathcal{D}}

\newcommand{\Grid}{\mathcal{G}}
\newcommand{\Comp}{\mathcal{A}}
\newcommand{\App}{a}
\newcommand{\GridDens}{\mathcal{G}_{\Dens}}
\newcommand{\GridApp}{\mathcal{G}_{\App}}
\newcommand{\GridAppMul}{\mathcal{G}^{\text{5D}}_{\App}}
\newcommand{\Vector}{\mathbf{v}}
\newcommand{\Matrix}{\mathbf{M}}
\newcommand{\AppVec}{\mathbf{b}}
\newcommand{\LightVec}{\mathbf{e}}
\newcommand{\AppMat}{\mathbf{B}}
\newcommand{\OuterP}{\circ}

\newcommand{\Loss}{\ell}
\newcommand{\LossRF}{\Loss_\text{RF}}
\newcommand{\LossPB}{\Loss_\text{PB}}
\newcommand{\LossBRDF}{\Loss_{\BRDFParam}}
\newcommand{\LossNormal}{\Loss_\Normal}
\newcommand{\LossDir}{\Loss_\Raydir}
\newcommand{\LossTensor}{\Loss_\Grid}

\newcommand{\LossW}{\alpha}
\newcommand{\LossWRF}{\LossW_\text{RF}}
\newcommand{\LossWPB}{\LossW_\text{PB}}
\newcommand{\LossWBRDF}{\LossW_{\BRDFParam}}
\newcommand{\LossWNormal}{\LossW_\Normal}
\newcommand{\LossWDir}{\LossW_\Raydir}
\newcommand{\LossWTensor}{\LossW_\Grid}

\begin{figure*}[t]
\centering
\vspace{-1mm}
\includegraphics[width=0.95\linewidth]{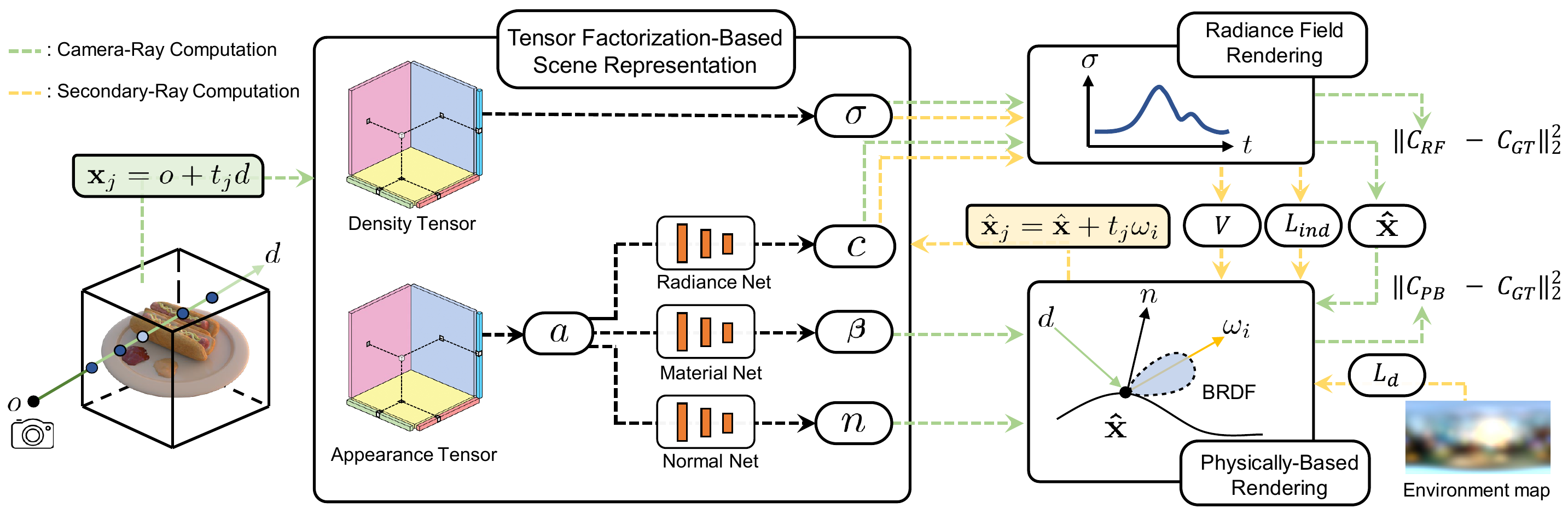}
\vspace{-2mm}
\caption{Overview. We propose a novel inverse rendering approach to reconstruct scene geometry, materials, and unknown natural illumination (as an environment map) from captured images.
We reconstruct a scene as a novel representation (Sec.~\ref{sec:representation}) that uses factorized tensors and multiple MLPs to regress volume density $\Dens$, view-dependent color $\Rad$, normals $\Normal$, and material properties (i.e. BRDF parameters) $\BRDFParam$, enabling both radiance field rendering and physically-based rendering (Sec.~\ref{sec:rendering}). 
In particular, we march a camera ray from camera origin $\Camorg$ in viewing direction $\Raydir$, sample points $\Pos_j$ along the ray, 
and apply radiance field rendering using the density and view-dependent colors regressed from our representation (Eqn.~\ref{equ:rfrendering}).
We also use the volume rendering weights to determine the surface point $\SurfP$ on the ray (Eqn.~\ref{equ:surf}), at which we perform physically based rendering using the normals and material properties (Eqn.~\ref{equ:pbrender}).
We compute accurate visibility $\Vis$ and indirect lighting $\Lindi$ using radiance field rendering by marching secondary rays from the surface point $\SurfP$ along sampled incoming light direction $\SurfDirIn$ (Sec.~\ref{sec:indirect}), enabling accurate physically-based rendering. 
We supervise both the radiance field rendering $\ColorRF$ and physically-based rendering $\ColorPB$ with the captured images in a per-scene optimization for joint scene reconstruction (Sec.~\ref{sec:recon}).
}
\vspace{-3mm}
\label{fig:pipeline}
\end{figure*}

\section{Related Works}

\boldstartspace{Neural scene representations.}
Neural representations \cite{xu2018deep,xu2019deep,thies2019deferred,lombardi2019neural,aliev2020neural,bi2020deepref,
mildenhall2021nerf},
as a promising alternative to traditional representations (like meshes,
volumes, and point clouds), have been revolutionizing 3D content generation and modeling.
Compared to traditional representations, such neural representations are more
flexible and can more accurately reproduce the geometry and appearance of
real-world scenes.  In particular, NeRF and many following neural field representations
\cite{mildenhall2021nerf,yariv2020multiview,xu2022point,muller2022instant,chen2022tensorf}
have been proposed and applied to enable high-fidelity rendering results on
novel view synthesis and many other applications \cite{park2021nerfies,chan2021pi,chan2021,li2021neural,zhang2022arf}.
Originally neural fields \cite{mildenhall2021nerf} are modelled in the form of MLPs, which have 
limited capacity and high computational costs. Recently, many 
works~\cite{muller2022instant, chen2022tensorf, sun2022direct} introduce more efficient 
neural scene representations that combine neural
feature maps/volumes with light-weight MLPs to reduce the computational cost and
accelerate the training and rendering. In this work, we adapt the efficient 
tensor-factorized neural representation, TensoRF~\cite{chen2022tensorf}, to achieve accurate and efficient inverse rendering.  

\boldstartspace{Inverse rendering.} 
Abundant works \cite{lawrence2004lightstage,hernandez2008multiview,goldman2009shape, xia2016recovering, bi2020deep, nam2018practical} 
have been proposed to infer the geometry and material 
properties of real-world objects from image collections. They typically 
represent the scene geometry using triangle meshes that are known or can be pre-reconstructed 
with depth sensors or multi-view stereo techniques~\cite{schoenberger2016mvs}. 
To reduce the ambiguities in the inverse rendering, they often require 
controlled lighting conditions~\cite{bi2020deep,nam2018practical}, 
or make use of learnt domain-specific priors~\cite{bi2020deep,dong2014appearance,barron2015shape,li2018learning}. 
In this work, we jointly estimate the geometry, materials and 
lighting from images captured under unknown lighting conditions with neural field representation that is more efficient and robust. 
While neural representations have recently been used for inverse rendering tasks,
they~\cite{boss2021nerd,NeuralPIL,zhang2021physg,bi2020neural,zhang2022iron,InvRenderer,zhang2021nerfactor,srinivasan2021nerv} 
are limited by the usage of computation-intensive MLPs. 
This inefficiency adds extra burden on inverse rendering when computing
secondary shading effects (like shadows), which requires to extensively sample secondary rays.
Therefore, previous methods often ignore secondary effects
\cite{boss2021nerd,NeuralPIL,zhang2021physg}, consider collocated flash lighting
\cite{bi2020deepref,bi2020neural,zhang2022iron}, or take extra costs to distill
these effects into additional MLP networks \cite{srinivasan2021nerv,zhang2021nerfactor,InvRenderer}.
In contrast, we base our model on the advanced TensoRF representation, utilizing
factorized tensors, to achieve fast reconstruction. 
Our TensoRF-based approach supports fast density and radiance evaluation,
enabling efficient online computation for secondary effects; this leads to more
accurate shadow and indirect lighting modeling during reconstruction, further
benefiting our reconstruction. 
Moreover, in contrast to previous methods
\cite{srinivasan2021nerv,zhang2021nerfactor,zhang2021physg,InvRenderer} that can
only handle captures under a single lighting condition, our model is easily extended to
support multi-light capture by modeling an additional lighting dimension in the
tensor factors.
\vspace{-4mm}

\section{Method}
In this section, we present our tensor factorization-based inverse rendering framework, shown in Fig.~\ref{fig:pipeline}, which reconstructs scene geometry, materials, and illumination from multi-view input images under unknown lighting.
We leverage rendering with both neural radiance fields and physically-based light transport to model and reproduce scene appearance (Sec.~\ref{sec:rendering}).
We introduce a novel TensoRF-based scene representation that allows for both rendering methods in scene reconstruction (Sec.~\ref{sec:representation}).
Our representation not only enables accurate and efficient computation for visibility and indirect lighting (Sec.~\ref{sec:indirect}) but also supports capturing under multiple unknown lighting conditions (Sec.~\ref{sec:multilight}).
We simultaneously estimate all scene components in a joint optimization framework with rendering losses and regularization terms (Sec.~\ref{sec:recon}).

\subsection{Rendering}
\label{sec:rendering}
We apply ray marching to achieve differentiable radiance field rendering 
as done in NeRF\cite{mildenhall2021nerf}. 
We further determine the expected surface intersection point for each 
ray using the volume density, and perform physically-based rendering at the surface point with the predicted 
scene properties. 

\boldstartspace{Radiance field rendering.}
Given a camera ray $\mathbf{r}(t)=\mathbf{o}+t\mathbf{d}$ from ray origin $\Camorg$ in direction $\Raydir$, radiance field rendering samples $N$ points on the ray and compute the pixel color as
\begin{equation}
\begin{gathered}
    \ColorRF(\Camorg, \Raydir) =\sum_{\iiRay=1}^{N} \Trans_{\iiRay}(1-\exp (-\Dens_{\iiRay} \delta_{\iiRay})) \Rad_{\iiRay}\\ \Trans_{\iiRay}=\exp(-\sum_{\iiTran=1}^{\iiRay-1} \Dens_{\iiTran} \delta_{\iiTran})
\end{gathered} 
\label{equ:rfrendering}
\end{equation}

where $\Dens_\iiRay$, $\delta_j$,  $\Rad_\iiRay$ and $\Trans_\iiRay$ are volume density, step size, view-dependent radiance color, and volume transmittance at each sampled point $\mathbf{r}(t_\iiRay)$.

\boldstartspace{Physically-based rendering.}
We apply a physically-based parametric BRDF model \cite{disneyBRDF} $\BRDFunc$ 
and perform physically-based rendering using predicted geometry and material 
properties at surface points on the camera rays. Similar to previous methods \cite{mildenhall2021nerf,zhang2021nerfactor}, these surface points $\SurfP$ are naturally determined using the volume rendering weights and sampled points from Eqn.~\ref{equ:rfrendering}:
\begin{equation}
    \SurfP = \sum_{\iiRay=1}^{N} \VolW_{\iiRay} \Ray(t_\iiRay), \quad \VolW_{\iiRay}=\Trans_{\iiRay}(1-\exp (-\Dens_{\iiRay} \delta_{\iiRay}))\label{equ:surf}
\end{equation}

We leverage the classic surface rendering equation to compute a physically-based shading color with an integral over the upper hemisphere $\Omega$ at each surface point:
\begin{equation}
\label{equ:pbrender}
    \ColorPB(\SurfP, \SurfDirO) = \int_\Omega 
    \Lin(\SurfP, \SurfDirIn)
    \BRDFunc(\SurfP, \SurfDirIn, \SurfDirO, \BRDFParam)
    (\SurfDirIn \cdot \Normal) \mathrm{d} \SurfDirIn
\end{equation}
where $\Lin(\SurfP, \SurfDirIn)$ is the incident illumination coming from direction $\SurfDirIn$. $\BRDFParam$ and $\Normal$ represent the spatially-varying material parameters (albedo and roughness) of the BRDF and the surface normal at $\SurfP$.

In theory, accurately evaluating this integral in Eqn.~\ref{equ:pbrender} 
requires extensively sampling the lighting direction and computing 
the incident lighting $\Lin(\SurfP, \SurfDirIn)$ to account for shadowing and 
indirect illumination.  It requires additional ray marching along each 
lighting direction, which is known to be extremely computation-expensive.
Previous NeRF-based methods often simplify it by only considering direct illumination \cite{boss2021nerd,NeuralPIL} or using extra auxiliary MLPs to make approximations 
and avoid full ray marching \cite{srinivasan2021nerv,zhang2021nerfactor}.
Instead, we evaluate the integral more accurately by marching secondary rays online and computing the incident lighting $\Lin(\SurfP, \SurfDirIn)$ with accurate shadowing and indirect illumination (see Sec.~\ref{sec:indirect}).
This is made possible by our novel efficient TensoRF-based scene representation.

\subsection{TensoRF-Based Representation}
\label{sec:representation}
We now introduce our tensor factorization-based scene representation that simultaneously models volume density $\Dens$, view-dependent color $\Rad$, shading normal $\Normal$, and material properties $\BRDFParam$ (including diffuse albedo and specular roughness of the Disney BRDF \cite{disneyBRDF}). 
Our method can model the scene under both single or multiple lighting conditions. 
We first introduce the single-light setting in this section and will discuss the multi-light extension in Sec.~\ref{sec:multilight}.

At a high-level, radiance, geometry, and material properties can all be represented as a 3D field. We can use a feature volume and voxel feature decoding functions to extract information at any point of the 3D field. To compress the space of the feature volume and also regularize the learning process, TensoRF~\cite{chen2022tensorf} proposed to use a low-rank factorized tensor as the feature volume. In this work, we adopt the Vector-Matrix factorization proposed by TensoRF. In particular, we use two separate VM-factorized tensors, i.e. feature grids $\GridDens$ and $\GridApp$, to model volume density and appearance, respectively. The appearance tensor $\GridApp$ is followed by multiple
light-weight MLP decoders to regress various appearance properties.

\boldstartspace{Density tensor.} The density tensor $\GridDens$ is 3D and directly expresses volume density without any network. Specifically, the VM-factorized density tensor is expressed by the sum of vector-matrix outer products.
\begin{align}
    \Grid_{\Dens} &= \sum_{k} \Vector_{\Dens, k}^{X} \OuterP \Matrix_{\Dens,k}^{YZ} + \Vector_{\Dens,k}^{Y} \OuterP \Matrix_{\Dens,k}^{XZ} + \Vector_{\Dens,k}^{Z} \OuterP \Matrix_{\Dens,k}^{XY} \notag\\
                 &= \sum_{k} \sum_{m\in XYZ} \Vector_{\Dens,k}^{m} \OuterP \Matrix_{\Dens,k}^{\Tilde{m}}
    \label{eqn:densevmr}
\end{align}
Here, $\Vector_{\Dens, k}^{m}$ and $\Matrix_{\Dens, k}^{\Tilde{m}}$ represent the $k^{\text{th}}$ vector and matrix factors of their corresponding spatial axes $m$; for simplicity, $\Tilde{m}$ denotes the two axes orthogonal to $m$ (e.g. $\Tilde{X}$=$YZ$).

\boldstartspace{Appearance tensor.}
The appearance tensor $\GridApp$ is 4D, modeled by similar vector-matrix spatial factors and additional feature basis vectors $\AppVec$, expressing a multi-channel voxel feature grid: 
\begin{align}
    \GridApp = \sum_{k} \sum_{m\in XYZ} \Vector_{\App,k}^{m} \OuterP \Matrix_{\App,k}^{\Tilde{m}} \OuterP \AppVec^m_k
    \label{eqn:appvmr}
\end{align}

\begin{figure}[t]
\includegraphics[width=\linewidth]{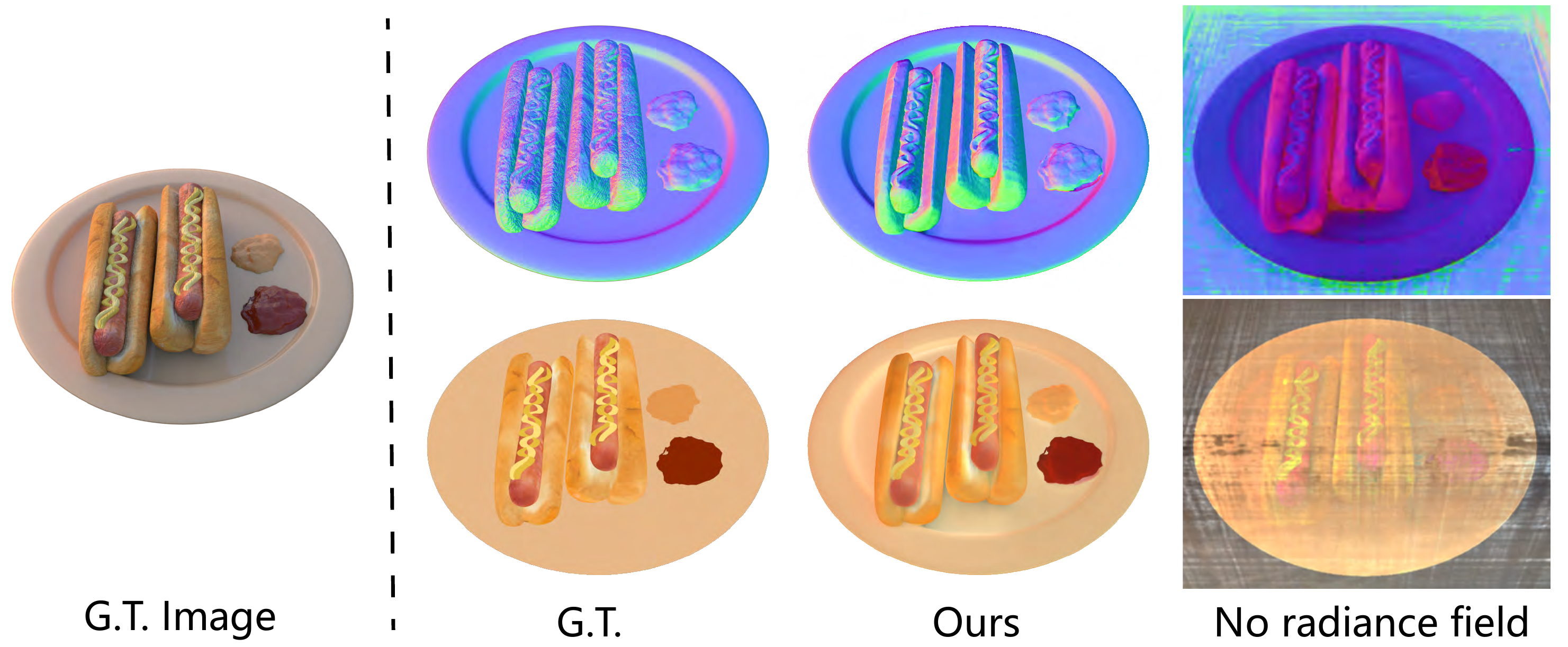}
    \vspace{-5mm}
    \caption{We compare normal and albedo reconstruction results between our joint reconstruction model and an ablated model without radiance field rendering during reconstruction. Radiance field reconstruction is crucial for us to achieve good reconstruction with a clean background and reasonable scene geometry.}
    \label{fig:rfcompare}
    \vspace{-5mm}
\end{figure}

\boldstartspace{Representing scene properties.}
We obtain density $\Dens$ directly by linearly interpolating $\GridDens$, and apply multiple MLP networks to decode appearance properties from interpolated appearance features $\GridApp$. 
This includes a radiance network $\MLP_\Rad$, a shading normal network $\MLP_\Normal$, and a material network $\MLP_{\BRDFParam}$.
Overall, our scene representation is expressed by
\begin{equation}
\begin{gathered}
    \Dens_\Pos = \GridDens(\Pos), \quad \App_\Pos = \GridApp(\Pos) \\
    \Rad_\Pos, \Normal_\Pos, \BRDFParam_\Pos = \MLP_\Rad(\App_\Pos),\MLP_\Normal(\App_\Pos),\MLP_{\BRDFParam}(\App_\Pos)
    \label{eqn:decoding}
    \end{gathered}
\end{equation}
Here, $\Pos$ denotes an arbitrary 3D location, $\Dens_\Pos = \GridDens(\Pos)$ and $\App_\Pos = \GridApp(\Pos)$ are the linearly interpolated density and multi-channel appearance features, computed by linearly interpolating the spatial vector and matrix factors (please refer to the TensoRF paper \cite{chen2022tensorf} for the details of feature computation and interpolation). 

In particular, volume density $\Dens$ and shading normals $\Normal$ both express scene geometry, which describes global shapes and high-frequency geometric details, respectively. 
View-dependent color $\Rad$ and physical shading properties (normal $\Normal$ and material parameters $\BRDFParam$) duplicatively model the scene appearance, determining the colors in the scene.
We bind shading normal with volume density using a regularization term (see details in Sec.~\ref{sec:recon}), correlating the scene geometry estimation and appearance reasoning.

In essence, our TensoRF-based scene representation provides scene geometry and appearance properties that are required for both radiance field rendering (Eqn.~\ref{equ:rfrendering}) and physically-based rendering (Eqn.~\ref{equ:pbrender}) described in Sec.~\ref{sec:rendering}.
\emph{This represents a scene as a radiance field and a physically-based rendering model jointly.}
We let the two sub-representations share the same neural features in our tensors, allowing their learning processes to benefit each other.
While the physically-based model is our main focus and achieves inverse rendering, modeling the radiance field can facilitate the volume density reconstruction and also regularize the appearance features to be meaningful since it has a shorter gradient path.
As a result, modeling the radiance field is crucial and necessary for us to achieve high-quality physically-based scene reconstruction as shown in Fig.~\ref{fig:rfcompare}.
In addition, the radiance field can naturally provide indirect illumination for physically-based rendering, enabling more accurate physical model reconstruction.

\begin{figure}[t]
\includegraphics[width=\linewidth]{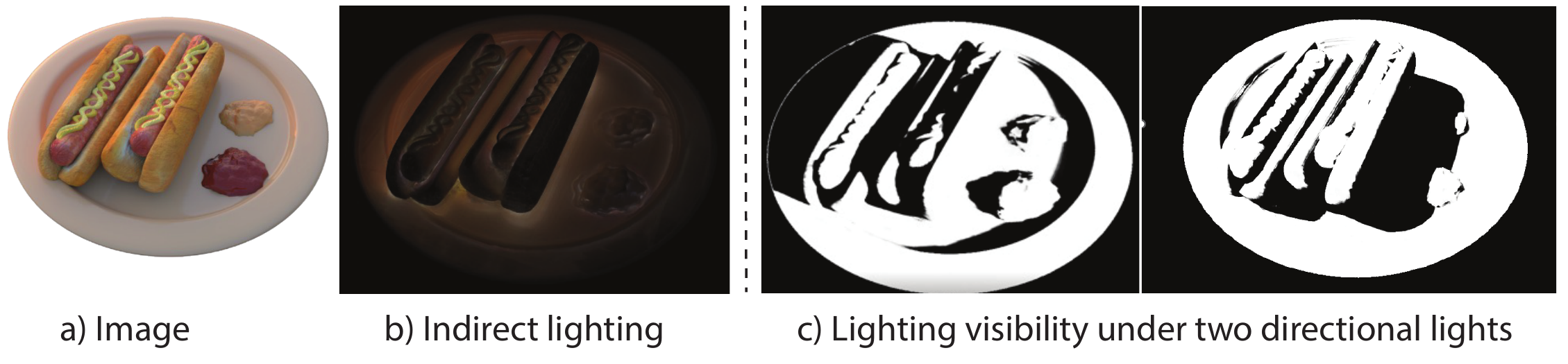}
    \vspace{-5mm}
    \caption{We show our computed indirect illumination (b) of the full rendered image (a) and our lighting visibility (c) under two different directional lights.}
    \label{fig:indirect}
    \vspace{-4mm}
\end{figure}

\subsection{Illumination and Visibility}
\label{sec:indirect}
Our TensoRF-based scene representation is highly efficient for optimization and evaluation.
Especially, our volume density can be computed by simple tensor interpolation without using any MLPs, leading to highly efficient computation of volume transmittance and rendering weights (Eqn.~\ref{equ:rfrendering},\ref{equ:surf}).
This allows us to compute accurate incident illumination $\Lin$ (as defined in Eqn.~\ref{equ:pbrender}) that accounts for secondary shading effects with ray marching as shown in Fig.~\ref{fig:indirect}. In particular, the incident illumination is computed by
\begin{equation}
\label{Equation:incoming_light}
\Lin(\SurfP, \SurfDirIn) =  \Vis(\SurfP, \SurfDirIn) \Ldirect (\SurfDirIn) + \Lindi(\SurfP, \SurfDirIn)
\end{equation}
where $\Vis$ is the light visibility function, $\Ldirect$ is the direct illumination, and $\Lindi$ is the indirect illumination. 

\boldstartspace{Direct illumination.} We assume the unknown natural environment to be distant from the captured object and represent the global illumination as an environment map, parameterized by a mixture of spherical Gaussians (SG), which represents the direct illumination. Note that in contrast to previous methods \cite{InvRenderer,zhang2021physg} that use SG for computing the integral of the rendering equation  with a closed-form approximation, 
we use SG only for its compact parameterized representation and compute the integral numerically by sampling secondary rays and performing ray marching, leading to more accurate lighting visibility and indirect lighting. 

\boldstartspace{Visibility and indirect illumination.}
We make use of the jointly-trained radiance fields to model secondary shading effects.
More specifically, the indirect illumination arriving at $\SurfP$ from $\SurfDirIn$ is inherently explained 
by the radiance color $\ColorRF$ along the ray $\Ray_i(t)=\SurfP+t\SurfDirIn$.
The visibility function is exactly modeled by the transmittance function in volume rendering. 
Therefore, the light visibility and indirect illumination term in Eqn.~\ref{Equation:incoming_light} 
can be calculated as:
\begin{equation}
\Vis(\SurfP, \SurfDirIn) = \Trans(\SurfP, \SurfDirIn), \quad \Lindi(\SurfP, \SurfDirIn) = \ColorRF(\SurfP, \SurfDirIn)
\end{equation}
Here $\Trans(\SurfP, \SurfDirIn)$ represents the volume transmittance of the final point sampled on the ray $\Ray_i(t)=\SurfP+t\SurfDirIn$.

\boldstartspace{Second-bounce ray marching.}
To make our physically-based rendering $\ColorPB$ accurate, we compute the rendering integral (Eqn.~\ref{equ:pbrender}) via Monte Carlo integration by marching multiple rays 
from each surface point $\SurfP$ 
with stratified sampling when training. \
For each ray, we obtain direct illumination from the SGs and compute visibility $\Trans(\SurfP, \SurfDirIn)$ and indirect illumination $\ColorRF(\SurfP, \SurfDirIn)$ using Eqn.~\ref{equ:rfrendering} directly.
This second-bounce ray marching is known to be expensive for previous NeRF-based methods, which either requires extremely high computational resources (128 TPUs) \cite{srinivasan2021nerv} or utilizes long (several days of) offline pre-computation \cite{zhang2021nerfactor}; both require training extra MLPs, which takes excessive computational costs and is unable to achieve high accuracy.  
We instead perform second-bounce ray marching online, achieving higher accuracy (see Fig.~\ref{fig:lightingshadow}), which is affordable, thanks to our highly efficient tensor factorization-based representation.

\subsection{Multi-Light Representation}
\label{sec:multilight}
We have discussed our scene representation under a single unknown lighting condition in Sec.~\ref{sec:representation}.
Our method can be easily extended to support capturing under multiple unknown lighting conditions, 
owing to the generality and extensibility of tensor factorization.
We achieve this by adding an additional lighting dimension with vector factors $\LightVec$ -- where the length of each vector $\LightVec$ equals the number of lighting conditions --  into the factorized appearance tensor $\GridApp$, leading to a 5D tensor that expresses scene appearance under different lighting conditions. We denote the 5D multi-light appearance tensor as $\GridAppMul$, represented by the factorization:
\begin{align}
    \GridAppMul = \sum_{k} \sum_{m\in XYZ} \Vector_{\App,k}^{m} \OuterP \Matrix_{\App,k}^{\Tilde{m}} \OuterP \LightVec^m_k \OuterP \AppVec^m_k
    \label{eqn:appvmrmulti}
\end{align}

Note that, it is only the view-dependent colors that require being modeled separately under different lighting, while the physical scene properties -- including volume density, normals, and material properties -- are inherently shared across multiple lighting conditions. Therefore, we decode view-dependent color using the neural features per light and decode other shading properties using the mean features along the lighting dimension: 
\begin{equation}
\begin{gathered}
    \Bar{\App}_{\Pos} = \frac{\sum_l \App_{\Pos,l}}{P}, \quad \App_{\Pos,l} = \GridAppMul(\Pos, l) \\
    \Rad_\Pos, \Normal_\Pos, \BRDFParam_\Pos = \MLP_\Rad(\App_{\Pos,l}),\MLP_\Normal(\Bar{\App}_{\Pos}),\MLP_{\BRDFParam}(\Bar{\App}_{\Pos})
    \label{eqn:decodingmulti}
    \end{gathered}
\end{equation}
where $l$ is the light index, $P$ is the number of lighting conditions, and $\Bar{\App}_{\Pos}$ is the mean feature. 

By simply adding additional vectors in the factorized appearance tensor, our method allows us to efficiently reconstruct and query the radiance under different illumination conditions, thereby still allowing for 
 indirect illumination computation in the multi-light setting with the method discussed in Sec.~\ref{sec:indirect}.
The multi-light input can provide useful photometric cues and reduce ambiguity in material prediction, 
and therefore leads to more accurate reconstruction of geometry and material estimation (see Tab.~\ref{tab:main_res}). 

\subsection{Joint Reconstruction and Training Losses}
\label{sec:recon}
For both single- and multi- light settings, we jointly reconstruct the scene geometry and appearance properties that are modeled by our tensor factorization-based representation, through an end-to-end per-scene optimization. 

\boldstartspace{Rendering losses.}
We supervise the rendered colors from both the radiance field rendering $\ColorRF$ and the physically-based rendering $\ColorPB$ with the ground-truth colors $\ColorGT$ from the captured images and include two
rendering loss terms:
\begin{equation}
    \LossRF = \|\ColorRF-\ColorGT\|^2_2, \quad \LossPB = \|\ColorPB-\ColorGT\|^2_2
\end{equation}

\boldstart{Normal regularization.}
Our shading normals $\Normal$ are regressed from an MLP to express high-frequency surface variations and used in physically-based rendering.
However, since the entire system is highly ill-conditioned, only supervising  
the network output with rendering loss is prone to overfitting issues and 
produces incorrect normal predictions. 
On the other hand, many previous NeRF-based method \cite{boss2021nerd, rudnev2022nerfosr, srinivasan2021nerv, zhang2021nerfactor} use the negative direction of volume density gradients $\Normal_\Dens =-\frac{\nabla_{\Pos} \Dens}{\left\|\nabla_{\Pos} \Dens\right\|}$ as normals for shading computation.
Such derived normals are better aligned with the surface but can be noisy and lack fine details.
Inspired by the rent work Ref-NeRF \cite{verbin2022refnerf}, we correlate our shading normal with the density-derived normal using a loss
\begin{equation}
    \LossNormal = \sum_{\iiRay} \VolW_{\iiRay} \| \Normal_\iiRay -\Normal_{\Dens,\iiRay} \|^2_2 
    \vspace{-3mm}
\end{equation}
This regularization term back-propagates gradients to both the appearance tensor (through the normal network) and the density tensor, further correlating our entire scene geometry and appearance reasoning, leading to better normal reconstruction. Note that while Ref-NeRF uses the same normal regularization, their normals are used in MLP-based shading computation; in contrast, our normals are used in a physically-based rendering process, allowing for more meaningful geometry reasoning and leading to more accurate normal reconstruction.

\begin{figure*}[t!]
\vspace{-7mm}
\includegraphics[width=\linewidth]{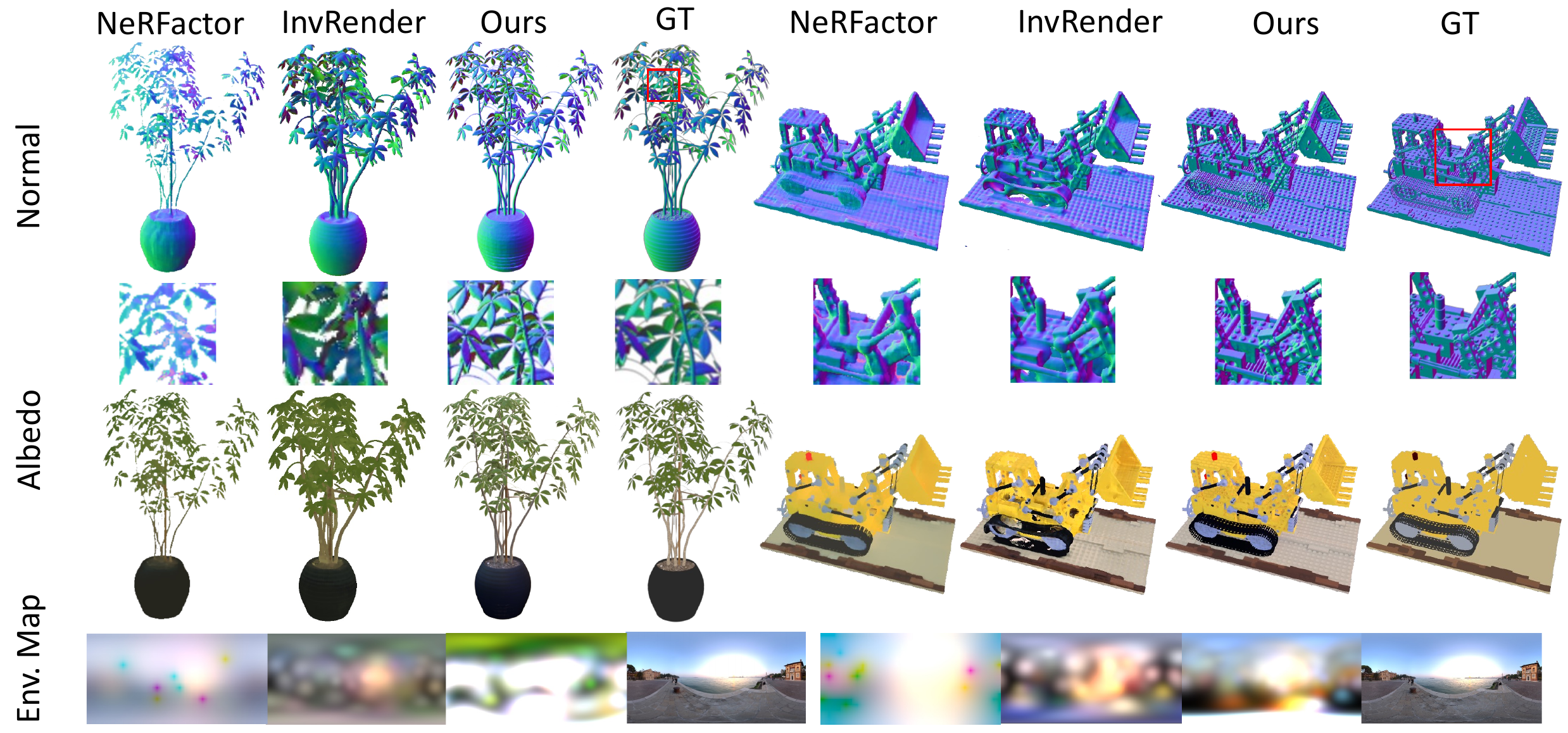}
    \vspace{-7mm}
    \caption{
    Visual comparison against baseline methods. Our method produces 
    inverse rendering results of higher quality with more detailed normals and more accurate albedo, thus leading to more photo-realistic relighting results. 
    }
    \vspace{-4mm}
    \label{fig:main_res}
\end{figure*}
\begin{table*}[ht]
    \centering
    \resizebox{\linewidth}{!}{
        \begin{tabular}{@{}c c c c ccc c ccc c ccc c c@{}}
            \toprule
            \multirow{2}{*}{Method} & & 
            \multicolumn{1}{c}{Normal} & & \multicolumn{3}{c}{Albedo} & & \multicolumn{3}{c}{Novel View Synthesis} & & \multicolumn{3}{c}{Relighting} 
            & &  \multirow{2}{*}{Average Runtime}  \\ \cline{3-3} \cline{5-7} \cline{9-11} \cline{13-15}
            & & MAE $\downarrow$ & & PSNR $\uparrow$ & SSIM $\uparrow$ & LPIPS $\downarrow$ & & PSNR $\uparrow$ & SSIM $\uparrow$ & LPIPS $\downarrow$ & & PSNR $\uparrow$ & SSIM $\uparrow$ & LPIPS $\downarrow$ \\ \hline
            
            NeRFactor & & 6.314 & & 25.125 & 0.940 & 0.109 & & 24.679 & 0.922 & 0.120 & & 23.383 & 0.908 & 0.131 
            & & \cellcolor{lightyellow}$>$ 100 hrs\\
            
            InvRender & & 5.074 & & 27.341 & 0.933 & 0.100 & & 27.367 & 0.934 & 0.089 & & 23.973 & 0.901 & 0.101 
            & & \cellcolor{yellow}15 hrs \\ \hline

            Ours in 25 minutes & & 
            \cellcolor{lightyellow}4.876 & & 
            \cellcolor{lightyellow}28.210 & \cellcolor{lightyellow}0.947 & \cellcolor{lightyellow}0.091 & & \cellcolor{lightyellow}32.350 & \cellcolor{lightyellow}0.964 & \cellcolor{lightyellow}0.061 & & \cellcolor{lightyellow}27.431 & \cellcolor{lightyellow}0.935 & \cellcolor{lightyellow}0.094 & & 
            \cellcolor{tablered}25 mins  \\ 
            
            Ours & & \cellcolor{yellow} 4.100 & & \cellcolor{yellow}29.275 & 
            \cellcolor{yellow}0.950 & 
            \cellcolor{yellow}0.085 & & 
            \cellcolor{tablered}35.088 & \cellcolor{tablered}0.976 & \cellcolor{tablered}0.040 & & 
            \cellcolor{yellow}28.580 & \cellcolor{yellow}0.944 & \cellcolor{yellow}0.081 
            & & \cellcolor{orange}5 hrs \\ \hline

            Ours w/ rotated multi-light & & 
            \cellcolor{orange}3.602 & & 
            \cellcolor{tablered}29.672 & \cellcolor{tablered}0.956 & \cellcolor{tablered}0.079 & & \cellcolor{orange}34.395 & \cellcolor{orange}0.974 & \cellcolor{orange}0.043 & & \cellcolor{orange}28.955 & \cellcolor{tablered}0.949 & \cellcolor{tablered}0.077 & & \cellcolor{orange}5 hrs  \\ 
            
            Ours w/ general multi-light & & \cellcolor{tablered}3.551 & & \cellcolor{orange}29.326 & \cellcolor{orange}0.951 & \cellcolor{orange}0.084 & & \cellcolor{yellow}34.223 & \cellcolor{yellow}0.973 & \cellcolor{yellow}0.045 & & \cellcolor{tablered}29.008 & \cellcolor{orange}0.947 &\cellcolor{orange} 0.078 & & \cellcolor{orange}5 hrs  \\ 
            
            \bottomrule
        \end{tabular}
    }
    \vspace{-3mm}
    \caption{Quantitative comparisons on the synthetic dataset. Our (single-light) results have 
    significantly outperformed the baseline methods by producing more accurate normal and albedo, 
    thus achieving more realistic novel view synthesis and relighting results. 
    Our method can further take images captured under different lighting conditions, and boost the performance in inverse rendering. (
    We scale each RGB channel of all albedo results by a global scalar, as done in NeRFactor \cite{zhang2021nerfactor}.
    For a fair comparison, all novel view synthesis results are generated with physically-based rendering, though our radiance field rendering has better quality.)
    }
    \vspace{-6mm}
    \label{tab:main_res}
\end{table*}

\boldstartspace{Final loss.}
Following TensoRF, we apply additional $\ell_1$-regularization on all tensor factors, denoted as $\LossTensor$.
We also apply a smoothness regularization term on BRDF parameters $\BRDFParam$ to enhance their spatial consistency. A loss term penalizing back-facing normals is used in addition. Please refer to the supplementary materials for the details of these regularization terms.
We apply all these losses to supervise and regularize our whole system to jointly reconstruct the scene, optimizing our scene representation (with all tensor factors and MLPs), as well as the SG parameters of the environment map, with a final loss
\begin{equation}
    \Loss = \LossWRF \LossRF + \LossWPB \LossPB + \LossWBRDF \LossBRDF + \LossWNormal \LossNormal +\LossWDir \LossDir +\LossWTensor \LossTensor
\end{equation}

\section{Experiments}
We now evaluate our model on various challenging scenes.
We  make comparisons against previous state-of-the-art
methods and also present an ablation study to verify the effectiveness of our design choices.

\boldstartspace{Datasets.}
We perform experiments on four complex synthetic scenes, including three 
blender scenes from~\cite{mildenhall2021nerf} and one from the Stanford 
3D scanning repository~\cite{curless1996volumetric}. 
We re-render these scenes to obtain their ground-truth images, as well as BRDF parameters and normal maps. We also render two types of multi-light data: rotated multi-light and general multi-light.
We also perform experiments on the original NeRF-synthetic dataset and 4 captured real data. Please refer to the supplementary for experimental results of those extra data, more details about our datasets, and more explanation about our multi-light setting.

\boldstartspace{Comparisons with previous methods.} We compare our model with previous state-of-the-art neural field-based inverse rendering methods, NeRFactor~\cite{zhang2021nerfactor} and InvRender~\cite{InvRenderer}, on these scenes using images captured under a single unknown lighting condition. We also compare with our model trained under multi-light settings.  
Table~\ref{tab:main_res} shows the accuracy of the estimated 
albedo, normal, and relighting results using metrics including
PSNR, SSIM and LPIPS~\cite{zhang2018unreasonable} averaged over the four scenes.   
We can see that our single-light method (using the same input as baselines) outperforms both previous methods significantly 
on all metrics, demonstrating the superiority of our method. 
We also include a visual comparison in Fig.~\ref{fig:main_res}, showing that our method predicts more accurate albedo and normal that 
are closer to the ground truth compared to baseline methods, thus generating 
more realistic relighting results.
In particular, our results produce normals that are of much higher quality and more faithfully reflect the geometry variations, while the baseline methods produce over-smooth results that lack shape details. 
\begin{figure}[t]
\vspace{-2mm}
\includegraphics[width=1\linewidth]{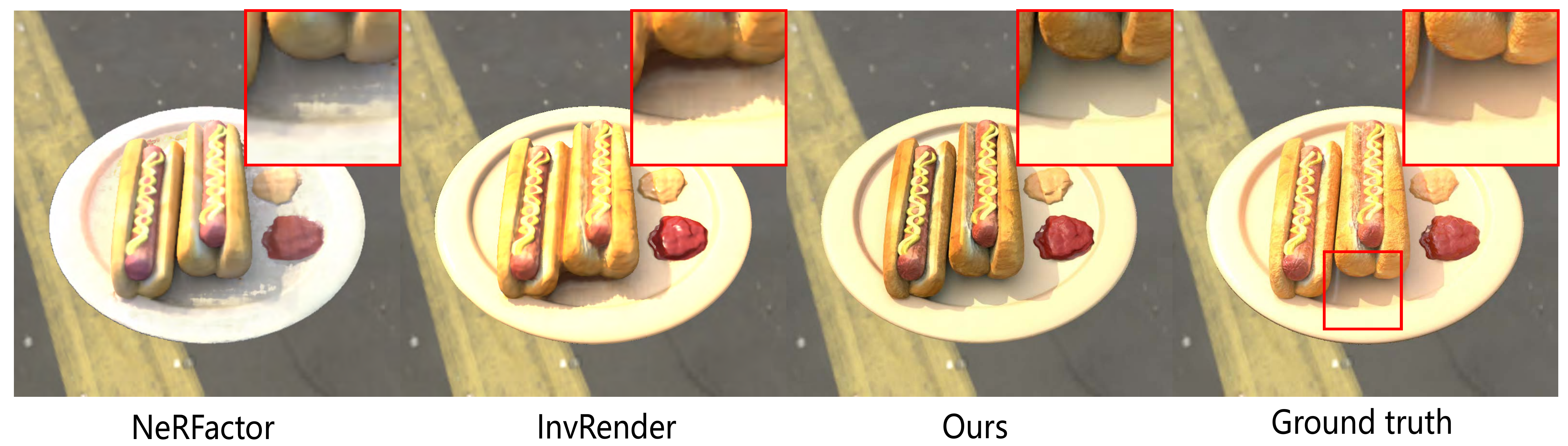}
    \vspace{-7mm}
    \caption{We compare our relighting results with previous methods. Note that our approach recovers more accurate shadows thanks to our second-bounce ray marching.}
    \label{fig:lightingshadow}
    \vspace{-7mm}
\end{figure}
Although NeRFactor's albedo result on the lego scene looks closer to the ground truth than our results. We claim that this is because that NeRFactor uses a high-weight BRDF smoothness loss, which damages its reconstruction quality of other components. In the supplementary, we showcase that we can achieve a similar albedo result by increasing the BRDF smoothness weight.
In particular, NeRFactor approximates the visibility function in a distilled MLP from a pre-trained NeRF and completely ignores the indirect illumination.
While InvRender considers both visibility and indirect illumination in their pre-computation, its spherical Gaussian-based shading computation can only achieve limited accuracy. 
Both methods do inverse rendering in a second stage after pre-training NeRFs.
In contrast, our tensor-factorized representation achieves a single-stage joint reconstruction and efficiently performs explicit second-bounce ray marching for more accurate shadowing (see Fig.~\ref{fig:lightingshadow}) and indirect lighting, thus leading to significantly higher reconstruction quality.

\begin{figure}[t]
\includegraphics[width=1\linewidth]{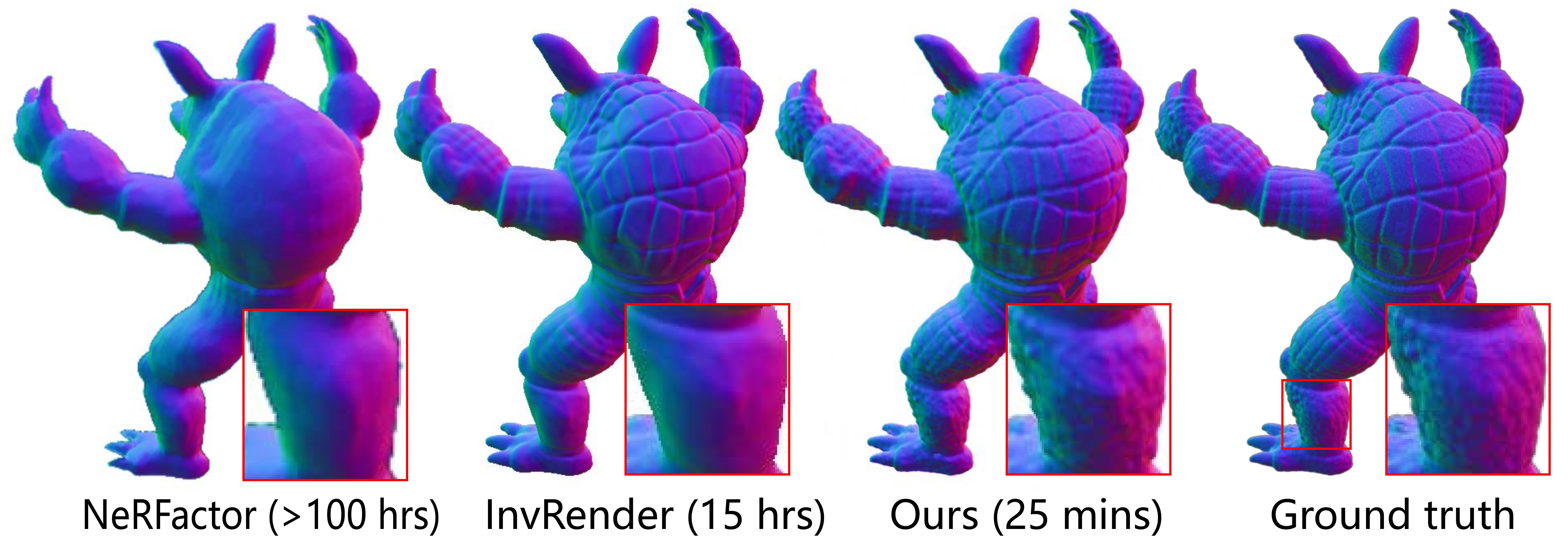}
    \vspace{-7mm}
    \caption{We compare geometry reconstruction results of our model taking only 25 minutes of optimization, with previous methods, taking 15 and $>100$ hours. Our approach even recovers more high-frequency details with substantially less reconstruction time.  }
    \label{fig:onehour}
    \vspace{-6mm}
\end{figure}

Meanwhile, owing to our more efficient representation, our high reconstruction quality is achieved with the fastest reconstruction speed, as reflected by the reconstruction time. NeRF-based method NeRFactor takes days to compute because of visibility pre-computation. InvRender is faster than NeRFactor, due to its SDF-based rendering~\cite{yariv2020multiview} and SG-based closed-form rendering integral computation; however, its SDF-based reconstruction fails on challenging scenes like the Lego shown in Fig.~\ref{fig:main_res} and the SG-based integration leads to inaccurate secondary shading effects as mentioned. 
On the other hand, our approach leverages ray marching-based radiance rendering in the reconstruction, robustly producing high-quality results on all testing scenes with accurate secondary effects, while still being faster than InvRender.
In fact, while our method takes 5 hours to finish its full training for its best performance, it can achieve good quality in a much shorter training period (25 minutes). As shown in ~\cref{tab:main_res} and Fig.~\ref{fig:onehour}, our approach can achieve high-quality geometry reconstruction in only 25 minutes and outperform previous methods that trained for tens of or even hundreds of hours.

\boldstartspace{Multi-light results.} In addition, our framework can also effectively leverage the additional input in a multi-light capture and further boost the accuracy of the inverse rendering performance without adding additional computation costs, as shown in Tab.~\ref{tab:main_res}, while the MLP-based baseline methods cannot be trivially extended to support such setups in an efficient manner. We also found that multi-light settings can greatly improve geometry reconstruction and help solving the color ambiguity between lighting and materials. Please refer the supplementary for more results and analysis.

\boldstartspace{Indirect illumination and visibility.}
Our tensor-factorized representation allows us to sample secondary 
rays in an efficient way to account for computing lighting visibility and indirect
illumination.  As shown in Tab.~\ref{tab:ablation_indirect_light}, without including these terms, the model cannot accurately represent the secondary shading effects and tend to bake them in the albedo or normal, resulting in lower quality. 
Nonetheless, note that these ablated models of our method in fact already achieves superior quality compared to previous methods (that pre-computes NeRFs and do inverse rendering disjointly) shown in Tab.~\ref{tab:main_res}, demonstrating the effectiveness of our joint reconstruction framework. 
Our full method further achieves more accurate reconstruction because of more accurate light transport modeling.

\begin{table}[t]
    \centering
    \resizebox{\linewidth}{!}{
        \begin{tabular}{c|c|c|c}
            \toprule
            Method & Normal MAE $\downarrow$ &  Albedo PSNR $\uparrow$&  NVS PSNR $\uparrow$ \\ \hline 
            w/o visibility & 4.716 & 27.265 & 34.703 \\
            w/o indirect illum. & 4.186 & 29.003 & 34.910 \\
            w/ indirect illum. + visibility & \textbf{4.100} & \textbf{29.275} & \textbf{35.088} \\
            \bottomrule
        \end{tabular}
    }
    \vspace{-4mm}
    \caption{Our full method models more physically-accurate light transport  
    by accounting for lighting visibility and indirect illuminations, thus achieving much better accuracy in 
    inverse rendering.}
    \vspace{-4mm}
    \label{tab:ablation_indirect_light}
\end{table}

\section{Conclusion}
We present a novel inverse rendering approach that achieves efficient and high-quality scene reconstruction from multi-view images under unknown illumination.
Our approach models a scene as both a radiance field and a physically-based model with density, normals, lighting, and material properties.
By jointly reconstructing both models, we achieve high-quality geometry and reflectance reconstruction, enabling photo-realistic rendering and relighting.
Owing to the efficiency and generality of the tensor factorized representation, our framework allows for accurate computation for shadowing and indirect lighting effects, and also flexibly supports capturing under an arbitrary number of lighting conditions.
We demonstrate that our approach is able to achieve state-of-the-art inverse rendering results, outperforming previous neural methods in terms of both reconstruction quality and efficiency. 

{\small
\bibliographystyle{ieee_fullname}
\bibliography{egbib}
}

\end{document}


\title{TensoIR: Tensorial Inverse Rendering \\ Supplementary Material}
\maketitle

\appendix

\section{Overview}
In this supplementary material, we show more results of our method, including the detailed per-scene reconstruction results of the four synthetic scenes (Sec.~\ref{sec:perscene}) and additional reconstruction results on four complex real scenes (Sec.~\ref{sec:real}). 
We discuss the implementation details of our method and give an analysis of our design choices and the effects of different loss weights in Sec.~\ref{sec:impl}.
Then, in Sec.~\ref{sec:multilight}, we give more details about our setups on synthetic dataset generation and our multi-light capture, and provide 
an in-depth analysis of the multi-light results. 
Finally, we discuss the limitations of our methods in Sec.~\ref{sec:Limitations}.

\begin{table*}[t!]
    \centering
    \resizebox{\linewidth}{!}{
        \begin{tabular}{@{} c c c c c c ccc c ccc c ccc@{}}
            \toprule
            \multirow{2}{*}{Scene} & &
            \multirow{2}{*}{Method} & & 
            \multicolumn{1}{c}{Normal} & & \multicolumn{3}{c}{Albedo} & & \multicolumn{3}{c}{Novel View Synthesis} & & \multicolumn{3}{c}{Relighting}  \\ \cline{5-5} \cline{5-9} \cline{11-13} \cline{15-17}
            & & & & MAE $\downarrow$ & & PSNR $\uparrow$ & SSIM $\uparrow$ & LPIPS $\downarrow$ & & PSNR $\uparrow$ & SSIM $\uparrow$ & LPIPS $\downarrow$ & & PSNR $\uparrow$ & SSIM $\uparrow$ & LPIPS $\downarrow$ \\ \hline

            \multirow{4}{*}{Lego} & & NeRFactor & & 9.767 & & 25.444 & 0.937 & 0.112 & & 26.076 & 0.881 & 0.151 & & 23.246 & 0.865 & 0.156 \\
            
            & & InvRender & & 9.980 & & 21.435 & 0.882 & 0.160 & & 24.391 & 0.883 & 0.151 & & 20.117 & 0.832 & 0.171 \\ 

            & & Ours (25 min) & & 7.780 & & 26.000 & 0.910 & 0.138 & & 32.180 & 0.952 & 0.061 & & 26.935 & 0.912 & 0.114 \\
                        
            & & Ours & & 5.980 & & 25.240 & 0.900 & 0.145 & & 34.700 & 0.968 & 0.037 & & 27.596 & 0.922 & 0.095 \\
            
            & & Ours w/ three rotated lights & & 5.630 & & 25.640 & 0.909 & 0.141 & & 34.590 & 0.968 & 0.037 & & 27.705 & 0.928 & 0.088 \\

            & & Ours w/ three general lights & & 5.370 & & 25.560 & 0.905 & 0.146 & & 34.350 & 0.967 & 0.038 & & 27.517 & 0.922 & 0.091 \\ \hline

            \multirow{4}{*}{Hotdog} & & NeRFactor & & 5.579 & & 24.654 & 0.950 & 0.142 & & 24.498 & 0.940 & 0.141 & & 22.713 & 0.914 & 0.159 \\
            
            & & InvRender & & 3.708 & & 27.028 & 0.950 & 0.094 & & 31.832 & 0.952 & 0.089 & & 27.630 & 0.928 & 0.089 \\ 
            
            & & Ours (25 min) & & 4.330 & & 29.390 & 0.947 & 0.099 & & 34.920 & 0.967 & 0.068 & & 27.353 & 0.927 & 0.124 \\
                        
            & & Ours & & 4.050 & & 30.370 & 0.947 & 0.093 & & 36.820 & 0.976 & 0.045 & & 27.927 & 0.933 & 0.115 \\
            
            & & Ours w/ three rotated lights & & 3.240 & & 30.180 & 0.959 & 0.079 & & 35.310 & 0.972 & 0.051 & & 28.459 & 0.939 & 0.110 \\

            & & Ours w/ three general lights & & 3.220 & & 31.240 & 0.958 & 0.080 & & 35.670 & 0.973 & 0.048 & & 28.952 & 0.939 & 0.110 \\ \hline

            \multirow{4}{*}{Armadillo} & & NeRFactor & & 3.467 & & 28.001 & 0.946 & 0.096 & & 26.479 & 0.947 & 0.095 & & 26.887 & 0.944 & 0.102 \\
            
            & & InvRender & & 1.723 & & 35.573 & 0.959 & 0.076 & & 31.116 & 0.968 & 0.057 & & 27.814 & 0.949 & 0.069 \\ 
            
            & & Ours (25 min) & & 2.360 & & 31.860 & 0.983 & 0.068 & & 35.160 & 0.978 & 0.053 & & 32.358 & 0.968 & 0.056 \\
                        
            & & Ours & & 1.950 & & 34.360 & 0.989 & 0.059 & & 39.050 & 0.986 & 0.039 & & 34.504 & 0.975 & 0.045 \\
            
            & & Ours w/ three rotated lights & & 1.590 & & 34.960 & 0.990 & 0.058 & & 38.480 & 0.985 & 0.041 & & 34.889 & 0.977 & 0.042 \\

            & & Ours w/ three general lights & & 1.550 & & 34.270 & 0.989 & 0.057 & & 38.230 & 0.984 & 0.043 & & 34.941 & 0.977 & 0.043 \\ \hline
            
            \multirow{4}{*}{Ficus} & & NeRFactor & & 6.442 & & 22.402 & 0.928 & 0.085 & & 21.664 & 0.919 & 0.095 & & 20.684 & 0.907 & 0.107 \\
            
            & & InvRender & & 4.884 & & 25.335 & 0.942 & 0.072 & & 22.131 & 0.934 & 0.057 & & 20.330 & 0.895 & 0.073 \\ 
            
            & & Ours (25 min) & & 5.040 & & 25.590 & 0.948 & 0.059 & & 27.140 & 0.958 & 0.062 & & 23.076 & 0.935 & 0.083 \\
                        
            & & Ours & & 4.420 & & 27.130 & 0.964 & 0.044 & & 29.780 & 0.973 & 0.041 & & 24.296 & 0.947 & 0.068 \\
            
            & & Ours w/ three rotated lights & & 3.950 & & 27.910 & 0.968 & 0.038 & & 29.200 & 0.972 & 0.043 & & 24.765 & 0.951 & 0.067 \\

            & & Ours w/ three general lights & & 4.060 & & 26.220 & 0.952 & 0.054 & & 28.640 & 0.967 & 0.050 & & 24.622 & 0.949 & 0.068 \\
            
            \bottomrule
        \end{tabular}
    }
    \caption{Per-scene results on the synthetic datasets.}
    \label{tab:synthetic_results}
\end{table*}

\section{Per-Scene Results on the Synthetic Dataset}
\label{sec:perscene}

In \cref{tab:synthetic_results}, we provide the results for individual synthetic scenes mentioned in Sec. 4 of the main paper. 
Our method outperforms both baselines in all four scenes. 
\Cref{fig:synt_res1} and \cref{fig:synt_res2} show our recovered normal, albedo, roughness, and relighting results from both our single- and multi-light-models on the four synthetic scenes.

\section{Results on Real-World Captures}
\label{sec:real}
\begin{figure}[t]
    \centering
    \includegraphics[width=0.8\linewidth]{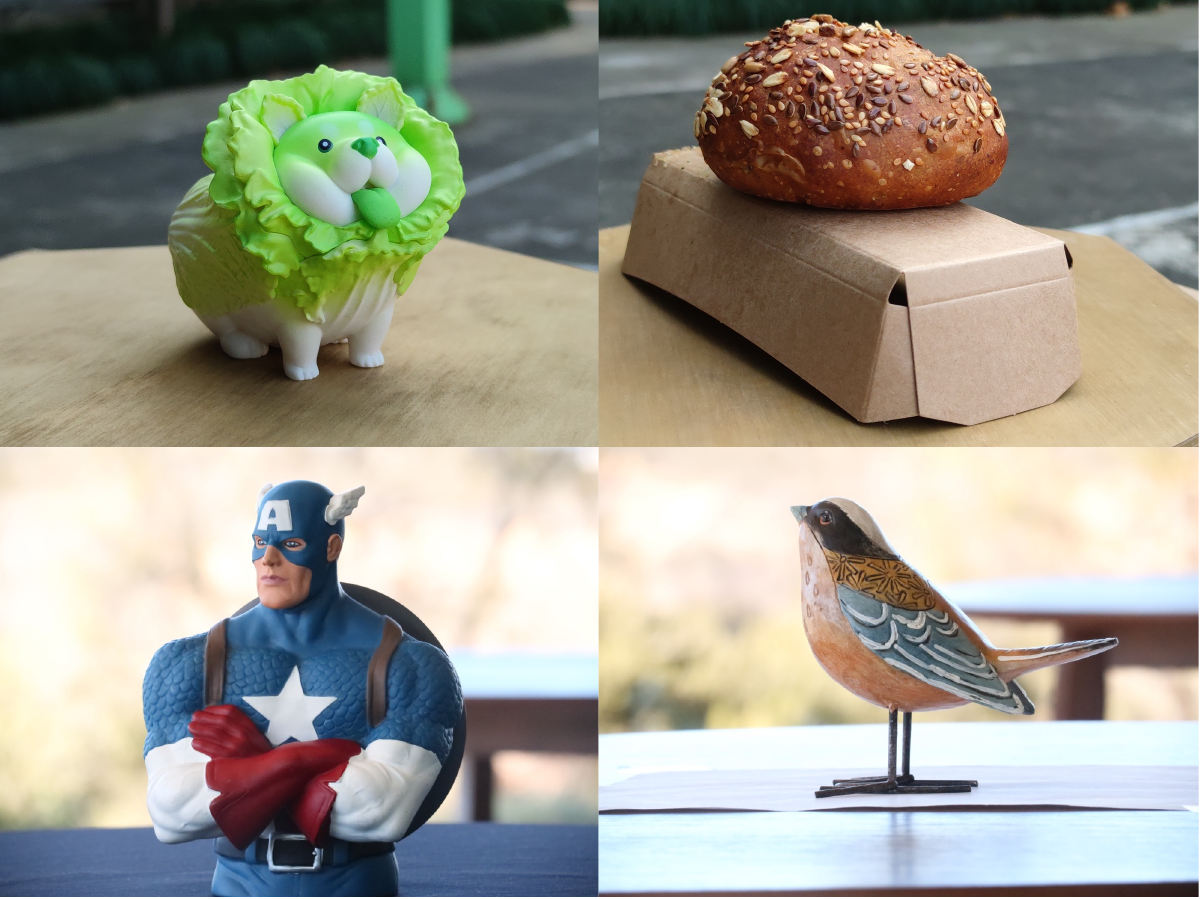}
    \caption{Four real objects we captured under natural lighting conditions. Please see Fig.~\ref{fig:realres} for their reconstruction results.}
    \label{fig:realobj}
\end{figure}

We capture 4 real objects (shown in \cref{fig:realobj}) under natural illumination in the wild to evaluate our method on real data. When capturing, we fix the camera parameters (exposure time, ISO, etc) and (roughly) uniformly take photos around the object. 
We use commercial software (picwish and remove.bg) to remove the background in each photo and use COLMAP to estimate the camera poses. \Cref{fig:realres} shows our reconstructed geometry, BRDF, and lighting on the real data. 
Note the quality of our reconstruction is affected by practical issues, such as the background removal quality, imperfect camera calibration, and non-static environment lighting (since there could be people passing by our in-the-wild setup). 
Nonetheless, our real-data results are still of very high quality. Please also see our video for more visual results.

\section{Implementation Details}
\label{sec:impl}

\boldstartspace{Representation details.}
As described in Sec.~3.2 in the main paper, we use a 3D density tensor $\GridDens$ and a 4D appearance tensor $\GridApp$ in our TensoRF-based scene representation; both tensors are factorized as multiple tensor components with vector and matrix factors. 
As in TensoRF, our model generally works well for any spatial resolutions of the feature grids and any number of tensor components; in general, higher solutions and more components lead to better reconstruction quality. 
For most cases, we use a spatial resolution of $300^3$; to achieve better details on scenes with complex thin structures (like Ficus), we use a resolution of $400^3$.
For all results, we use 48 tensor components (16 components per dimension) for the density tensor and 144 components (48 components per dimension) for the appearance tensor separately.
For decoding the multiple appearance properties, we design our MLP decoder networks all as a small two-layer MLP with 128 channels in each hidden layer and ReLU activation. In addition, the Radiance Net receives the appearance feature and viewing direction as input, and the Normal Net and BRDF Net will receive intrinsic feature and 3D location as input. Frequency encoding is applied on both directions and locations.

For our multi-light representation (in Sec.~3.4 of the paper), we leverage the mean appearance feature $\Bar{\App}_{\Pos}$ (Eqn.~8) for normal and reflectance decoding. 
In practice, this mean is computed with the means of lighting vectors $\LightVec$, averaged along the lighting dimension, without computing individual $\App_{\Pos,l}$ for lower costs, leveraging the linearity of Eqn.~9).

\boldstartspace{Training details.}
We run our model on a single RTX 2080 Ti GPU(11 GB memory) for all our results.
For fair comparisons, the baseline methods (NeRFactor and InvRender) are also re-run with the same GPU to test their run-time performance.
We train our full model using Adam optimizer; following TensoRF, we use initial learning rates of 0.02 and 0.001 for tensor factors and MLPs respectively.
We also perform coarse-to-fine reconstruction as done in TensoRF by linearly upsampling our spatial tensor factors (started from $N^3_0$=$128^3$ for all cases) multiple times during reconstruction until achieving the final spatial resolution ($N^3$=$300^3$ in most cases as mentioned). We upsample the vectors and matrices linearly and bilinearly at steps 10000, 20000, 30000, 40000 with the numbers of voxels interpolated between $N^3_0$ and $N^3$ linearly in logarithmic space.

The total training iteration is 80k and the average training time is 5 hours. 
The first 10k will be used to generate alphaMask, which is also used in the original TensoRF, to help skip empty space, so it only has radiance field rendering to compute image loss and only costs about 5 minutes. We do so because we find the alphaMask can greatly help to reduce the GPU memory cost of physically-based rendering: We find that if we directly perform physically-based rendering directly at the very beginning of the training process without generating the alphaMask, the training ray batch size can not be larger than 1024, otherwise, we would meet cuda-out-of-memory errors. And because we spend a few minutes generating a coarse alphaMask (which will be updated in the later training process), we can sample 4096 camera rays for each training batch. The number of points sampled per camera ray is determined by the grid resolution;  a grid size of $300^3$ leads to about 1000 points per ray. 

\boldstartspace{Ray marching details. }
During training, when computing the visibility and indirect lighting, we sample 512 secondary rays starting from each surface point with 96 points per ray, and half of the rays will be filtered according to surface normal orientation (only those rays whose directions lie in the upper hemisphere of the normal vector will be valid). The sampled rays' directions are generated by stratified sampling (We divide the environmental map into grids with equal 2D areas and generate a random direction inside each grid). 
Also, the visibility gradients and indirect lighting gradients are detached for GPU memory consideration.
While our secondary ray sample is coarser than primary (camera) ray sampling, we find this is enough to achieve accurate shadowing and indirect lighting computation.

During relighting, since we have testing ground-truth environmental maps as input, we use lighting-intensity importance sampling during ray sampling instead of stratified sampling, which means we will sample more rays near those directions that have stronger lighting. 

\boldstartspace{Loss details.}
Our model is reconstructed with a combination of multiple loss terms as introduced in Sec.~3.5 and Eqn.~13 in the paper.
We now introduce the details of the BRDF smoothness term $\LossBRDF$ and normal back-facing term $\LossDir$ in Eqn.~13. In particular, we impose scale-invariant smoothness terms on our BRDF predictions (both roughness and albedo) to encourage their spatial coherence. 
For each sample point on the camera ray, we minimize the relative difference of its predicted  material properties from those of the randomly-sampled 
neighboring points, defined as: 

\begin{equation}
\LossBRDF = \sum_{ j, \ \Pos=\Ray(t_\iiRay)} \VolW_\iiRay \left\| \frac{\BRDFParam_\Pos-\BRDFParam_{\Pos+\boldsymbol{\xi}}}{\mathbf{max}(\BRDFParam_\Pos, \BRDFParam_{\Pos+\boldsymbol{\xi}})} \right\|_{2}^{2}
\label{equ:brdf_smoothness}
\end{equation}
where $\boldsymbol{\xi}$ is a small random translation vector generated from a normal distribution with zero mean and 0.01 variance, and 
$w_j$ is the volume rendering weights (as described in Eqn.~2 in the paper) to assign large weights for points around the object surface. 
This weight $w_j$ has also been used for other loss terms (including the normal regularization term $\LossNormal$ in Eqn.~12).
In addition, we also regularize the predicted normals by penalizing those that are near the surface and back-facing with the orientation loss introduced by Ref-NeRF:
\begin{equation}
    \LossDir = \sum_{\iiRay}  \VolW_{\iiRay} \max(0, \Normal_{\iiRay} \cdot \Raydir) 
\end{equation}
We also have a TV loss shortly in the process of generating alphaMask to help eliminate some small floaters.

We set the radiance field rendering loss weight $\LossWRF$ to be 1.0, physically-based rendering loss weight $\LossWPB$ to be 0.2, and BRDF smoothness regulation loss weight to be 0.001. 
The $\ell_1$-regularization on all tensor factors has the same loss weight as TensoRF. \ \emph{The weight $\LossWNormal$ for normals difference loss $\LossNormal$ (the loss that constrains the difference between the predicted normals from Normal Net and the derived normals from the density field) is crucial for the final reconstruction quality.} \ We find reasonable weights to lie in $ \left [4\times10^{-4}, 6\times10^{-3} \right]$.
Larger normals difference loss $\LossWNormal$ can help to prevent the Normal Net prediction from overfitting on input images but will at the same time damage the network's ability to predict high-frequency details.

\begin{figure}[t]
    \centering
    \includegraphics[width=1\linewidth]{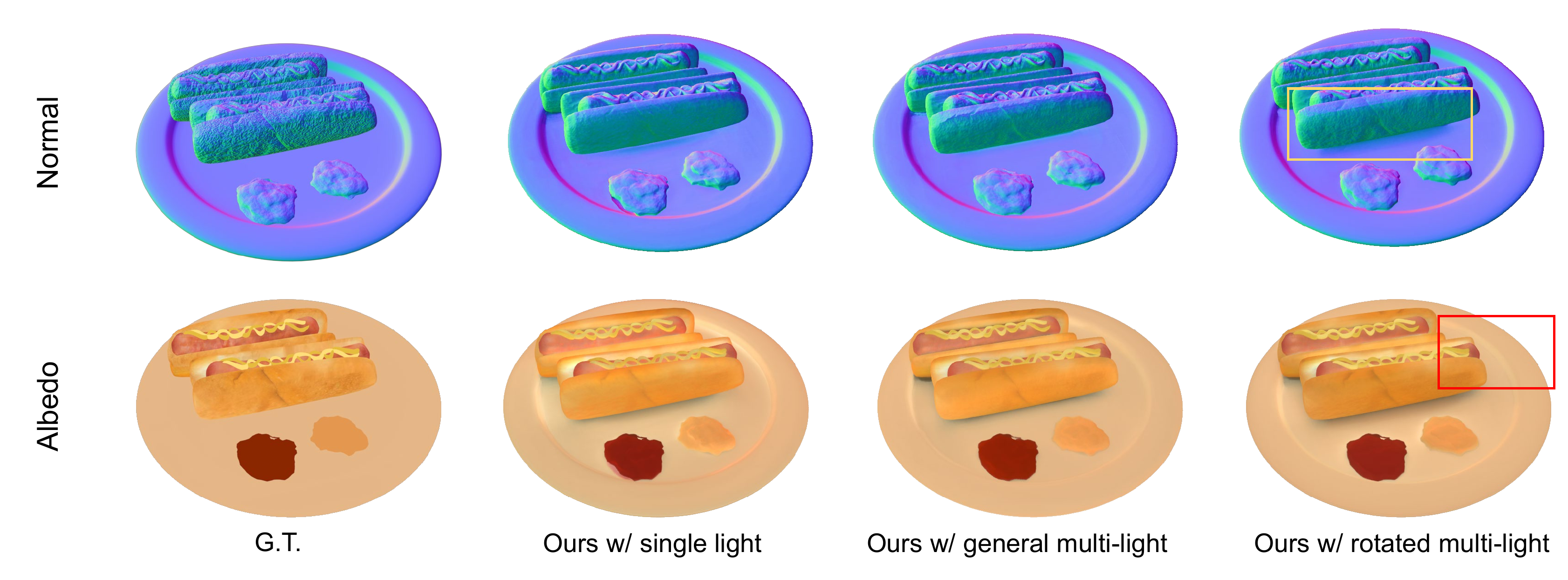}
    \caption{Comparison of single-light and multi-light results on synthetic data.}
    \label{fig:synt_multi_light_compare}
\end{figure}

\begin{figure}[t]
    \centering
    \includegraphics[width=1\linewidth]{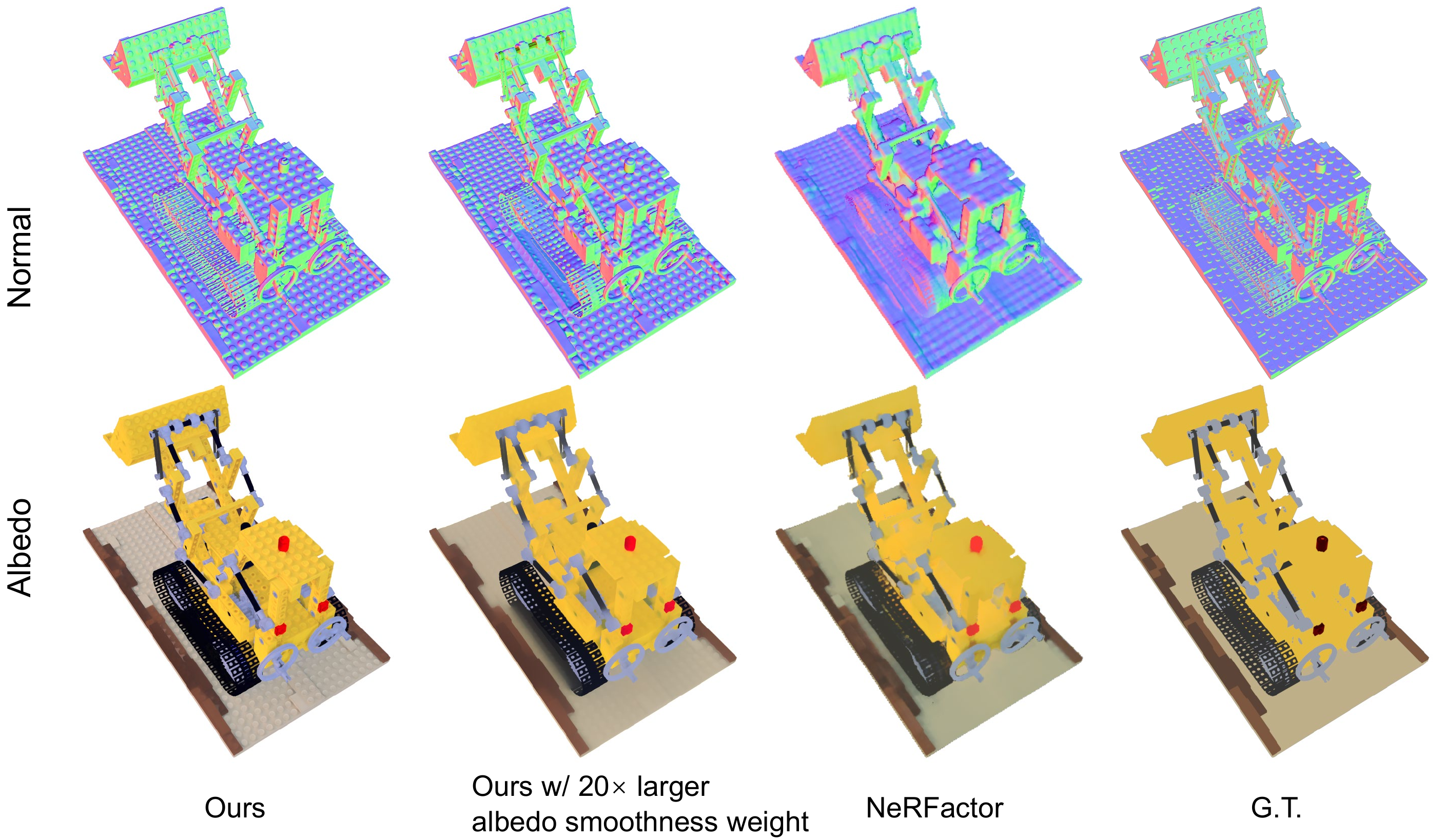}
    \caption{Comparison of NeRFactor's results, ours, and ours with larger albedo smoothness weight. 
    The above results show that with larger BRDF smoothness loss weight on lego scene, our method can get the similar smooth albedo recontruction result as NeRactor's result, but this will damage our geometry reconstruction quality (although our geometry result under this case is still better than NeRFactor).
    }
    \label{fig:smoothness_loss_on_lego}
\end{figure}

\boldstartspace{Effects of BRDF smoothness loss on lego's reconstruction.}
We give more analysis and explanations about the artifacts of our albedo reconstruction result on lego scene, which has been discussed partly in the main paper. As shown by Fig. 5 of the main paper, NeRFactor’s albedo result on the lego scene looks closer to the ground truth than
our results because its result looks smoother. In the main paper, we claim that this is because that NeRFactor
uses a high-weight BRDF smoothness loss, which helps it achieve smooth albedo reconstruction but damages its reconstruction quality of other components. As shown in ~\cref{fig:smoothness_loss_on_lego}, when making the loss weight of our albedo smoothness loss become 20 times larger, our albedo result will be smoother and closer to the ground truth, but this will damage our normal reconstruction quality (but still better than the results of our baselines). Therefore, to have better geometry reconstruction results and to make the loss weight of BRDF smoothness loss fixed across different scenes, we do not use extra larger BRDF smoothness loss weight for lego in our experiments.

\begin{figure}[t]
    \centering
    \includegraphics[width=0.7\linewidth]{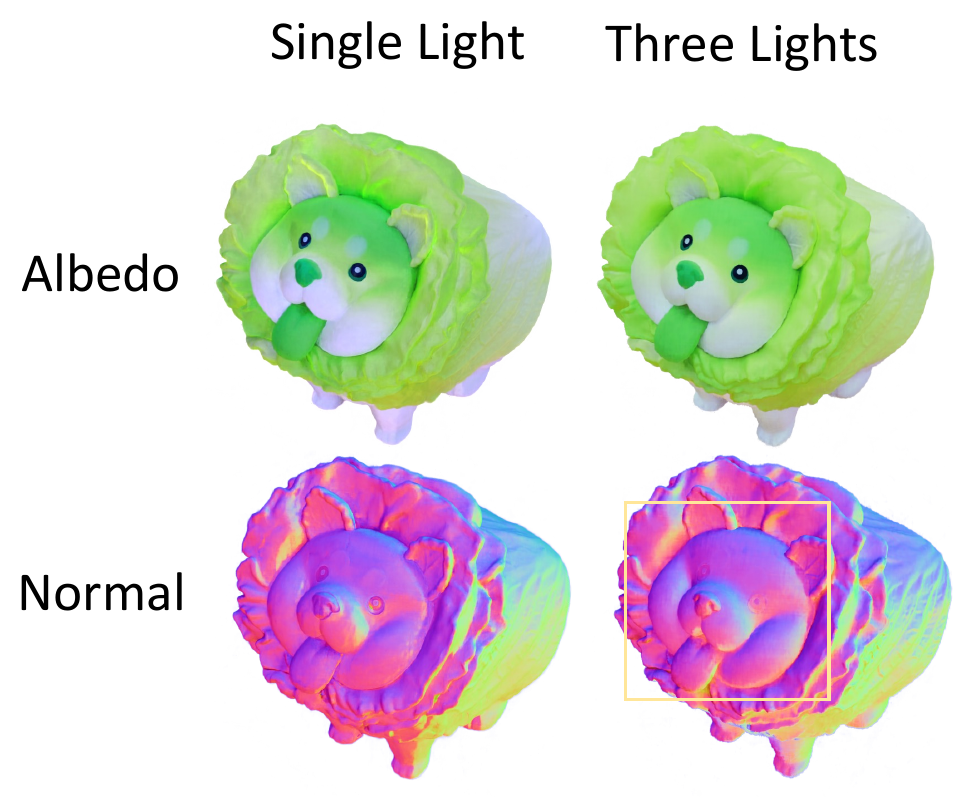}
    \caption{Comparison of single-light and multi-light (three-light) results on a real scene. The multi-light capture is achieved in a practical way by simply rotating the object three times under the same environment lighting.}
    \label{fig:real_multi_light_compare}
\end{figure}

\section{More Details and Analysis on Our Synthetic Dataset and Multi-Light Capture}
\label{sec:multilight}
\boldstartspace{More details about our synthetic dataset and multi-light settings.}
we perform experiments on four complex synthetic scenes, including three 
blender scenes (ficus, lego, and hotdog) from the original NeRF and one (armadillo) from the Stanford 
3D scanning repository.
All data are re-rendered by multiple high-resolution ($2048 \times 1024$) environment maps to obtain their ground-truth images ($800 \times 800$ resolution) for training and testing, as well as BRDF parameters and normal maps. We use the same camera settings as NeRFactor, so we have 100 training views and 200 test views. 

In the main paper, we discussed our results on two types of multi-light data: \textbf{rotated multi-light data} and 
\textbf{general multi-light data}.
\textbf{Rotated multi-light data}
is rendered under a rotated multi-light setting, in which the images are rendered from the same 100
views as in the single-light setting, but each view has 3 images
rendered by rotated environment maps. we rotate the same environment map along the azimuth for 0, 120, and 240 degrees, which can be done in practice by rotating the captured object (as done in ~\cref{fig:real_multi_light_compare}). And with known rotation degrees, our method can optimize shared environmental lighting across the rotated multi-light data. 
\textbf{General multi-light data}
is rendered under a general multi-light setting, in which we create three lighting conditions by rendering the objects with three unrelated environment maps, which will be optimized separately in the later training process.

Considering both multi-light settings above have more input training images than the single-light setting (the number of training views is the same, but multi-light settings above have more images per view), we introduce the third the multi-light setting here which is called \textbf{limited general multi-light setting} to evaluate whether the improvements of reconstruction quality under multi-light setting are due to the extra number of input images. It uses the same kind of data as the general multi-light setting, but for each view we will only randomly select one image as training input from the 3 images under different lighting conditions, which guarantees that the number of training images in this setting is the same as the single-light setting. As shown in ~\cref{tab:three_kinds_multi_light}, while using the same number of images, such a setting still achieves  
better performance in BRDF estimation and geometry reconstruction than the single-light setting,  which demonstrates the benefits of 
the multi-light input.

\begin{table}[t]
    \centering
    \resizebox{\linewidth}{!}{
        \fontsize{10pt}{12pt}\selectfont
        \begin{tabular}{c|c|c|c}
            \toprule
            Method & Normal MAE $\downarrow$ &  Albedo PSNR $\uparrow$&  NVS PSNR $\uparrow$ 
            \\ \hline 
            Ours w/ single-light& 4.100 & 29.275 & \textbf{35.258}\\            Ours w/ rotated multi-light & 3.602 & \textbf{29.672} & 34.395
            \\
            Ours w/ general multi-light & \textbf{3.551} & 29.326 & 34.395
            \\
            Ours w/ limited general multi-light & 3.670 & 29.320 & 34.100
            \\
            \bottomrule
        \end{tabular}
    }
    \caption{Quantitative comparisons of results on the synthetic dataset using single-light and multi-light input. }
\label{tab:three_kinds_multi_light}
\vspace{-5mm}
\end{table}

\begin{figure}[t]
    \centering
    \includegraphics[width=1.0\linewidth]{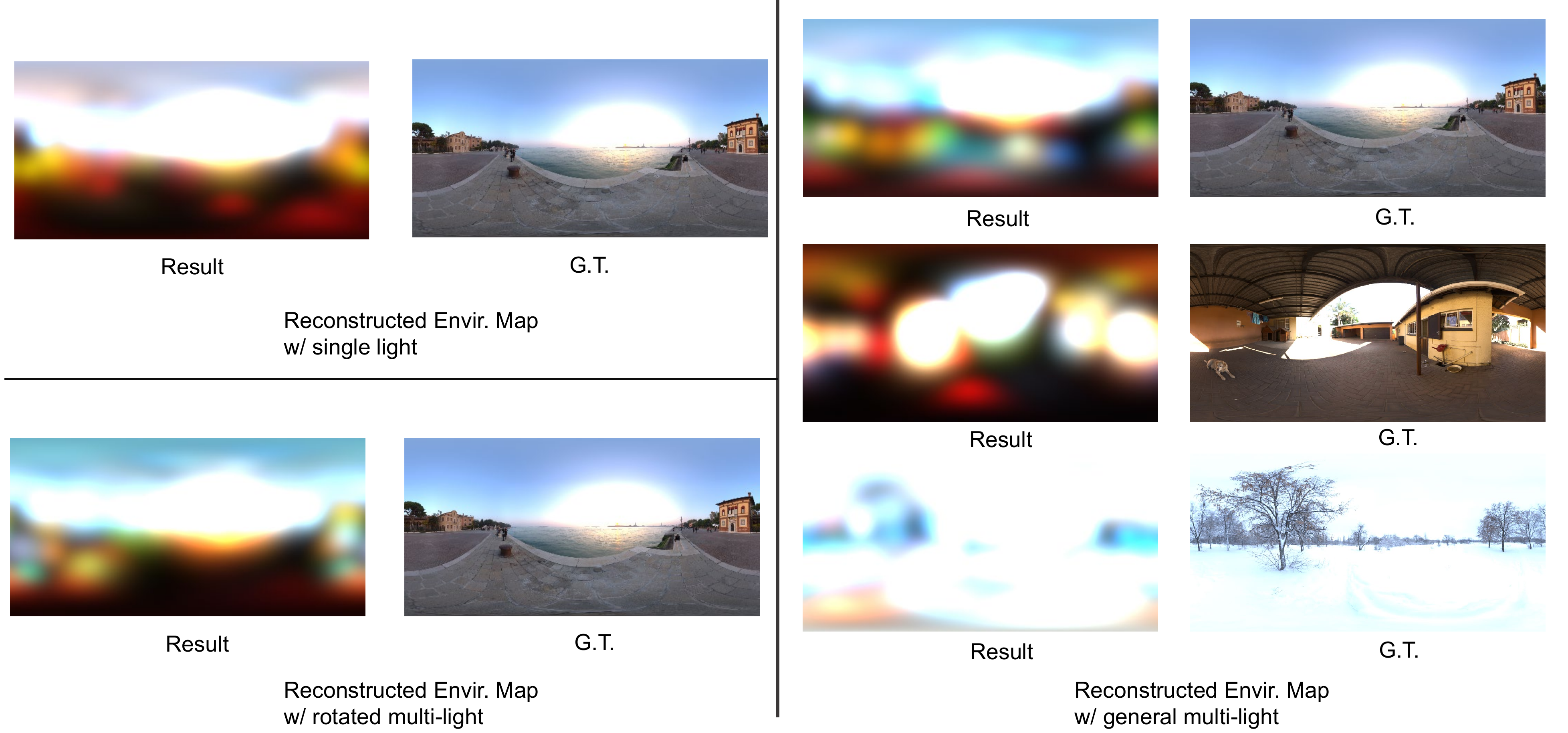}
    \caption{Comparison of reconstructed environment maps under single-light, rotated multi-light, and general multi-light settings. 
    Compared to single-light input, multi-light input enables more accurate reconstructions of the lighting. 
    }
    \label{fig:solving_color_ambiguity}
\end{figure}

\begin{figure*}[t]
    \centering
    \includegraphics[width=1.0\linewidth]{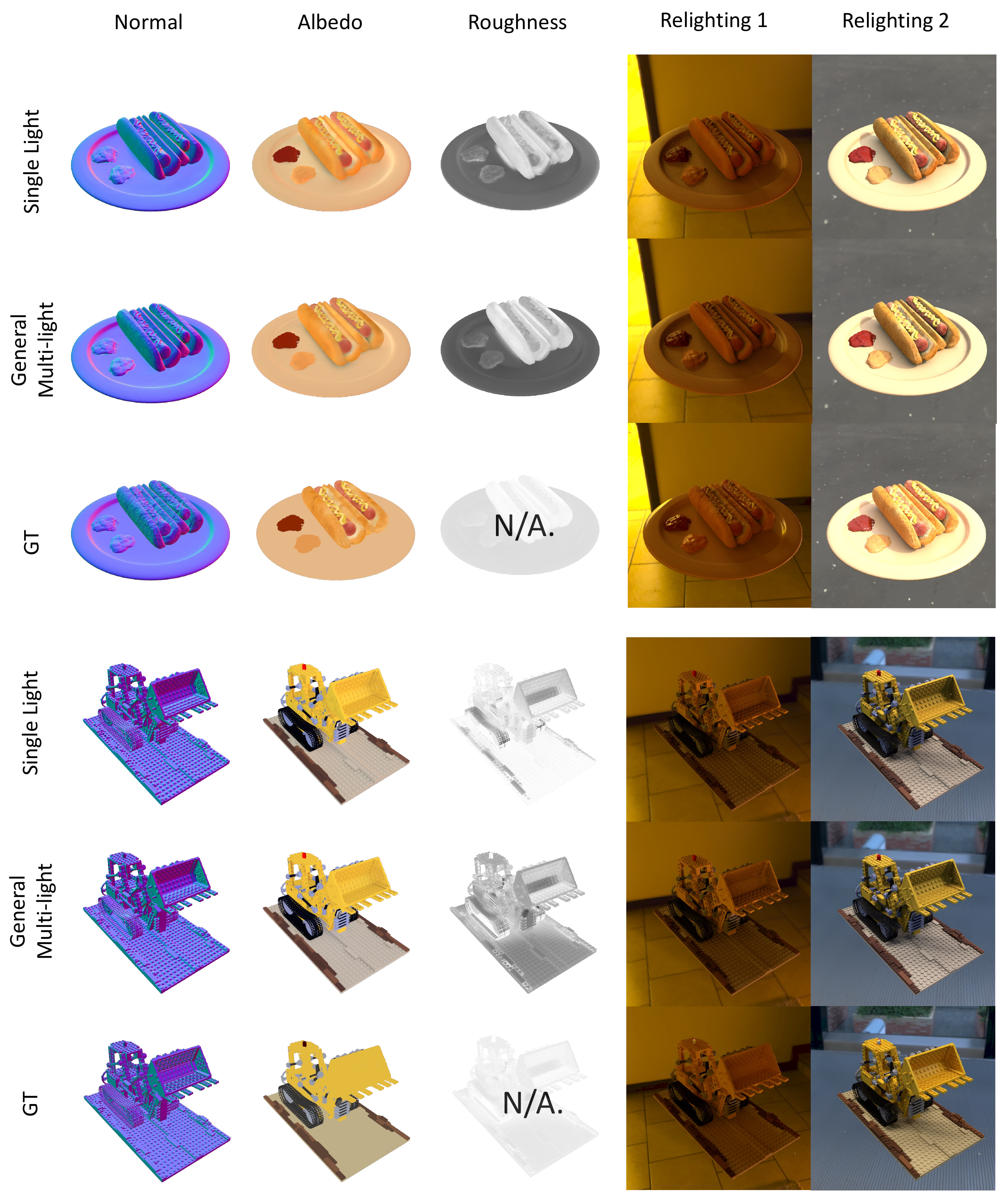}
    \caption{Our reconstructed normal, albeo, roughness and relighting results on hotdog and lego synthetic dataset. The G.T. roughness is marked as N/A, because the synthetic data is not rendered with the Disney BSDF model.}
    \label{fig:synt_res1}
\end{figure*}

\begin{figure*}[t]
    \centering
    \includegraphics[width=1.0\linewidth]{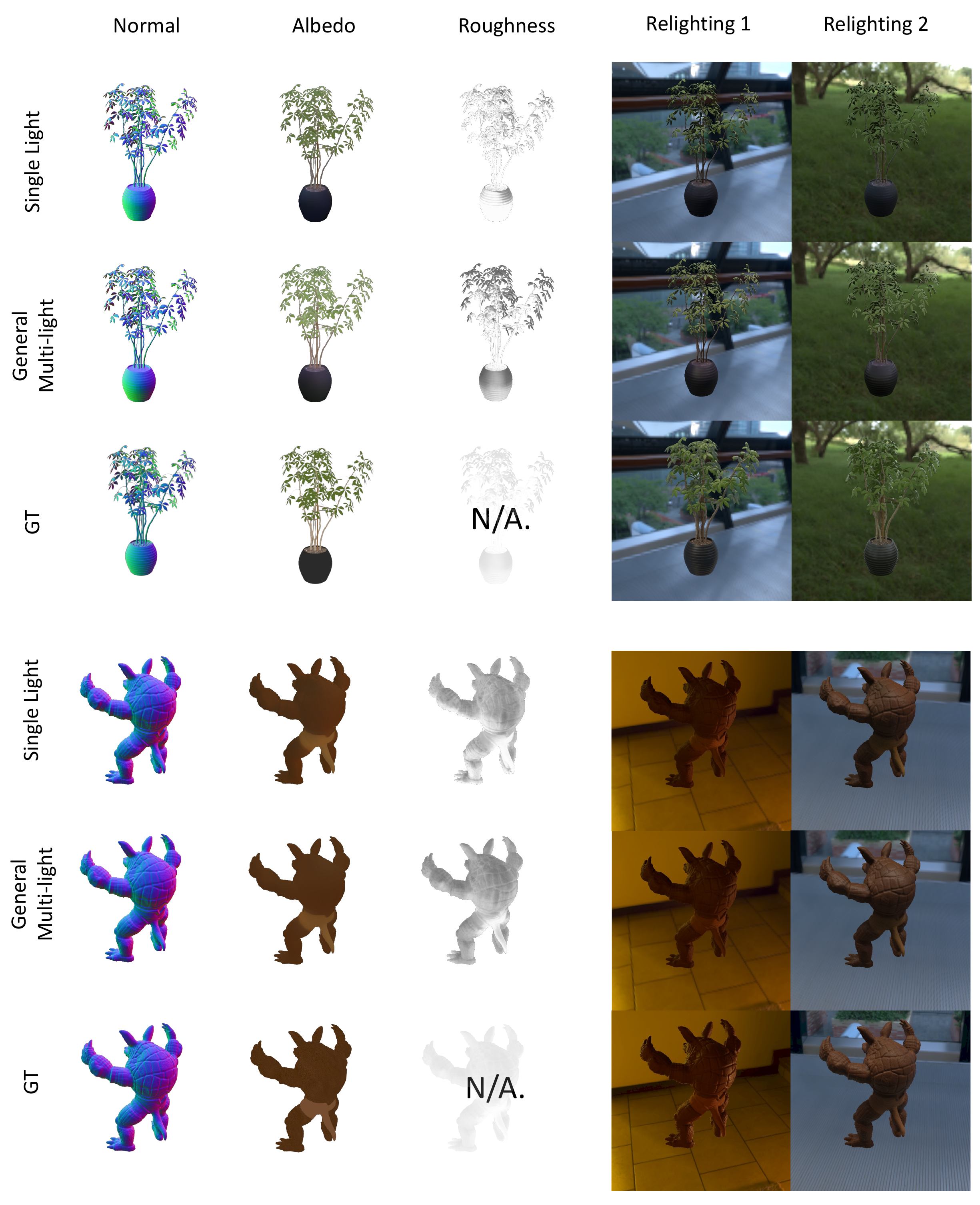}
    \caption{Our reconstructed normal, albeo, roughness and relighting results on ficus and armadillo synthetic datasets. The G.T. roughness is marked as N/A, because the synthetic data is not rendered with the Disney BSDF model.}
    \label{fig:synt_res2}
\end{figure*}

\begin{figure*}[t!]
    \centering
    \includegraphics[width=1\linewidth]{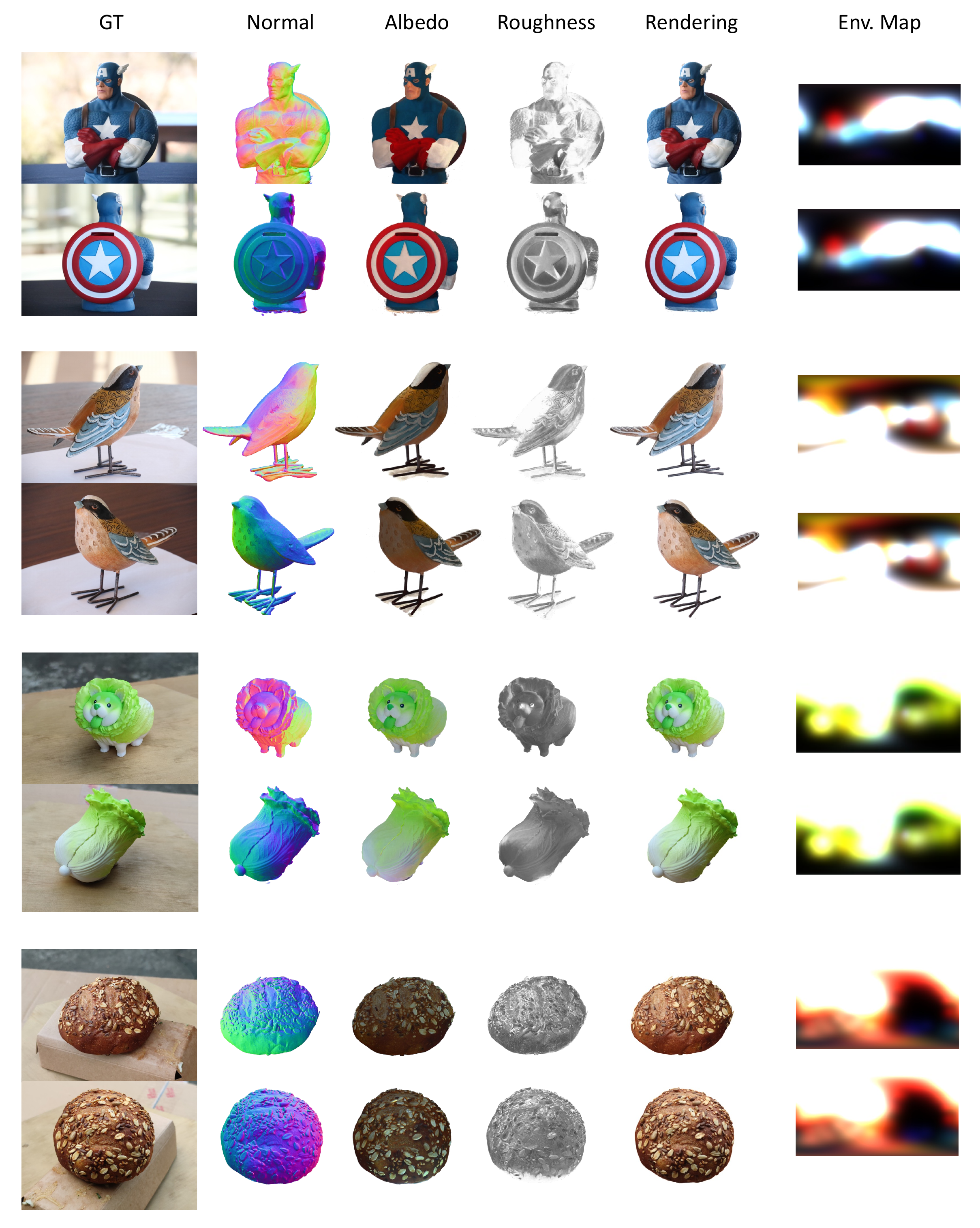}
    \caption{Decomposed albedo, roughness, normal, and lighting with our method on four real objects.}
    \label{fig:realres}
\end{figure*}

\boldstartspace{Analysis of results with multi-light captures.}
As shown in Tab.~\ref{tab:synthetic_results} (and also the main paper's Tab.~1),
our approach enables effective and efficient multi-light reconstruction, leading to better reconstruction accuracy.
While the single-light novel view synthesis (physically-based rendering) results are slightly better than our multi-light results, this is simply because we evaluate novel view synthesis under the same single lighting, which the single-light model is specifically trained on (and easier to overfit).
On the other hand, our multi-light model achieves better reconstruction and leads to much better rendering quality under novel lighting conditions (as shown by the relighting results).
We also show visual comparisons between our single-light and multi-light results on both synthetic and real scenes in Fig.~\ref{fig:synt_multi_light_compare} and Fig.~\ref{fig:real_multi_light_compare}.
Our multi-light reconstruction recovers more details in the normal maps and recovers better shading- and artifact-free albedo maps.
We also achieve rotated multi-light capture for the real data by simply rotating the object three times.
Our results in Fig.~\ref{fig:real_multi_light_compare} show that even such simple multi-light acquisition can already lead to high-quality reconstruction in practice (better than the single-light results), demonstrating the effectiveness of our multi-light reconstruction model. 
We also find that multi-light settings can help to solve the color ambiguity between environmental lighting and object materials. As shown by ~\cref{fig:solving_color_ambiguity}, the color of reconstructed environment map is closer to the ground truth under multi-light settings.

\section{Limitations}
\label{sec:Limitations}
\begin{figure*}[t!]
    \vspace{-3mm}
    \includegraphics[width=0.95\linewidth]{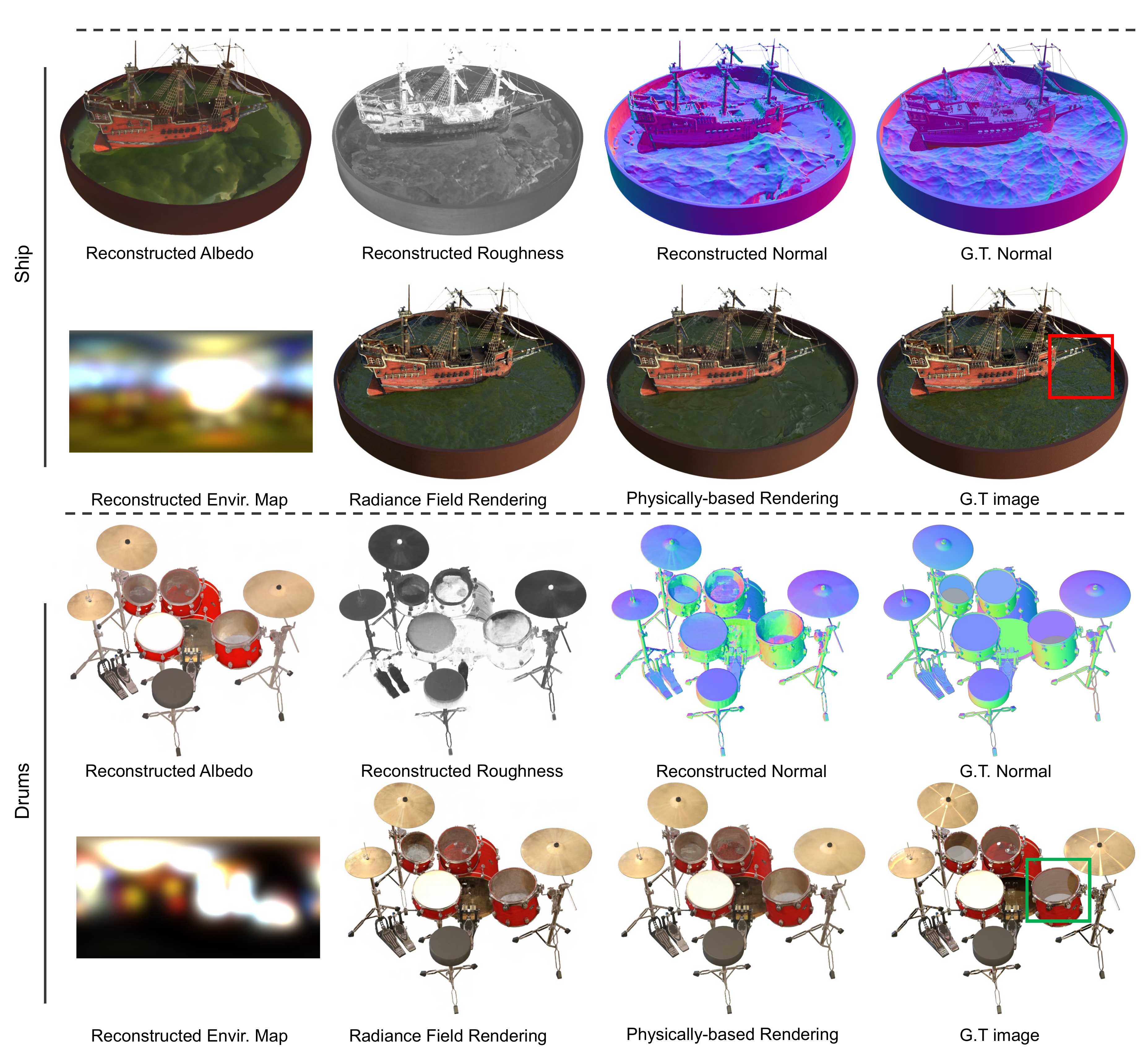}
    \caption{
    Limitations of our method.
    We run our method on challenging scenes from the original NeRF-Synthetic dataset to help analyze the limitations of our method: the Ship scene and Drums scene contain complex materials such as translucent water and transparent glasses, which 
    produce complex light transport effects that cannot be modeled by our existing surface-based rendering model.
    We also assume a fixed Fresnel term that is not suitable for modeling non-dielectric materials such as metals in the Drums scene. 
    Therefore, our method produces renderings with artifacts and incorrect geometries in these regions. 
    }
    \label{fig:nerf_synthetic}
\end{figure*}

We evaluate our method on complex scenes from the original NeRF-Synthetic dataset to help understand the limitations of our method.
In general, our approach has the following limitations: first, our physically-based rendering applies a surface-based rendering model, 
which means that we can not handle complex materials that can not be well modeled by this model, for example, translucent water that has 
strong reflection and refraction (see the red frame in ~\cref{fig:nerf_synthetic})
and transparent glass (see the green frame in ~\cref{fig:nerf_synthetic}).
Second, we assume the materials of the objects to be dielectric (non-conducting) and therefore fix the ${F_0}$ in the fresnel term of our simplified Disney BRDF to be 0.04, which means in theory, we can not well model non-dielectric materials like metals. Replacing the pure physically-based BRDF with a learning-based neural BRDF that has learned data prior about materials from training data can help overcome this limitation. We leave it to be future works.

